\documentclass[lettersize,journal]{IEEEtran}
\usepackage{amssymb}
\usepackage{pifont}
\newcommand{\cmark}{\ding{51}}%
\newcommand{\xmark}{\ding{55}}%
\usepackage{xcolor, colortbl}
\usepackage{times}
\usepackage{epsfig}
\usepackage{graphicx}
\usepackage{bm}
\usepackage{booktabs}
\usepackage{setspace}
\usepackage{float}
\usepackage{color}
\usepackage{multirow}
\usepackage{ amsmath, caption, subcaption, overpic, textpos}
\usepackage[numbers,sort&compress]{natbib}
\usepackage{bbm}

\usepackage[pagebackref=false, breaklinks=true, letterpaper=true, colorlinks,
            citecolor=citecolor, linkcolor=linkcolor, bookmarks=false]{hyperref}
\definecolor{citecolor}{HTML}{0071BC}
\definecolor{linkcolor}{HTML}{ED1C24}
\definecolor{Ins_enc}{HTML}{4672C4}

\newlength\savewidth\newcommand\shline{\noalign{\global\savewidth\arrayrulewidth
  \global\arrayrulewidth 1pt}\hline\noalign{\global\arrayrulewidth\savewidth}}
\newcommand{\tablestyle}[2]{\setlength{\tabcolsep}{#1}\renewcommand{\arraystretch}{#2}\centering\footnotesize}
\renewcommand{\paragraph}[1]{\vspace{1.25mm}\noindent\textbf{#1}}

\newcolumntype{x}[1]{>{\centering\arraybackslash}p{#1pt}}
\newcolumntype{y}[1]{>{\raggedright\arraybackslash}p{#1pt}}
\newcolumntype{z}[1]{>{\raggedleft\arraybackslash}p{#1pt}}

\newcommand{\app}{\raise.17ex\hbox{$\scriptstyle\sim$}}

\definecolor{deemph}{gray}{0.6}

\definecolor{baselinecolor}{gray}{.9}
\definecolor{deltacolor}{gray}{.45}
\newcommand{\baseline}[1]{\cellcolor{baselinecolor}{#1}}

\definecolor{airforceblue}{rgb}{1.00, 0.501, 0.01}
\definecolor{egreen}{rgb}{0, 0.69, 0.314}

\definecolor{cyan}{cmyk}{.3,0,0,0}
\usepackage[capitalize]{cleveref}
\crefname{section}{Sec.}{Secs.}
\Crefname{section}{Section}{Sections}
\Crefname{table}{Table}{Tables}
\crefname{table}{Tab.}{Tabs.}

\usepackage{soul}
	\definecolor{airforceblue}{rgb}{0.36, 0.54, 0.66}

\begin{document}
	
	\title{Joint Gaze-Location and Gaze-Object Detection}
	\author{Danyang Tu, Wei Shen,
            Wei Sun,
	        Xiongkuo Min, 
	        Guangtao Zhai
	}
	\maketitle
	\begin{abstract}
     This paper proposes an efficient and effective method for joint gaze location detection (GL-D) and gaze object detection (GO-D), \emph{i.e.}, gaze following detection. Current approaches frame GL-D and GO-D as two separate tasks, employing a multi-stage framework where human head crops must first be detected and then be fed into a subsequent GL-D sub-network, which is further followed by an additional object detector for GO-D. In contrast, we reframe the gaze following detection task as detecting human head locations and their gaze followings simultaneously, aiming at jointly detect human gaze location and gaze object in a unified and single-stage pipeline. To this end, we propose GTR, short for \underline{G}aze following detection \underline{TR}ansformer, streamlining the gaze following detection pipeline by eliminating all additional components, leading to the first unified paradigm that unites GL-D and GO-D in a fully end-to-end manner. GTR enables an iterative interaction between holistic semantics and human head features through a hierarchical structure, inferring the relations of salient objects and human gaze from the global image context and resulting in an impressive accuracy. Concretely, GTR achieves a 12.1 mAP gain ($\mathbf{25.1}\%$) on GazeFollowing and a 18.2 mAP gain ($\mathbf{43.3\%}$) on VideoAttentionTarget for GL-D, as well as  a 19 mAP improvement ($\mathbf{45.2\%}$) on GOO-Real for GO-D. Meanwhile, unlike existing systems detecting gaze following sequentially due to the need for a human head as input, GTR has the flexibility to comprehend any number of people's  gaze followings simultaneously, resulting in high efficiency. Specifically, GTR introduces over a $\times 9$ improvement in FPS and the relative gap becomes more pronounced as the human number grows.
	\end{abstract}
	
	\section{Introduction}
	\label{sec:intro}
    \IEEEPARstart{G}{aze} following is one of the extraordinary human abilities to precisely track others' gaze directions and identify their gaze targets. In real-world scenarios, the human visual system operates with speed and precision, allowing us to perform complex vision tasks with little conscious thoughts, such as effortlessly understanding the behavioural intentions of others. Similarly, fast and accurate gaze following detection algorithms will enable machines to better interpret human behaviors, thereby presenting a significantly potential in various human-centric vision tasks, such as saliency prediction~\cite{leifman}, human action detection~\cite{shapovalova2013action}, and human-object interaction detection~\cite{xu2019interact}, among others~\cite{zhang2020human}.
 
    Existing detection systems frame gaze following detection as two separate tasks, \emph{i.e.}, detecting \emph{where} each human is looking at (gaze location detection, GL-D)~\cite{chong, fang2021dual, tu2022end} and detecting \emph{what} each human is looking at (gaze object detection, GO-D)~\cite{tomas2021goo, wang2022gatector}. Specifically, the former aims to predict the gaze location of each human (\emph{i.e.}, a heatmap, the magnitude of different locations indicates the possibility that the location is the gaze point), while the latter further identify the category of the object being looked at (\emph{i.e.}, a bounding box with object label).

    \begin{figure}[t]
		\centering
		\includegraphics[width=\linewidth]{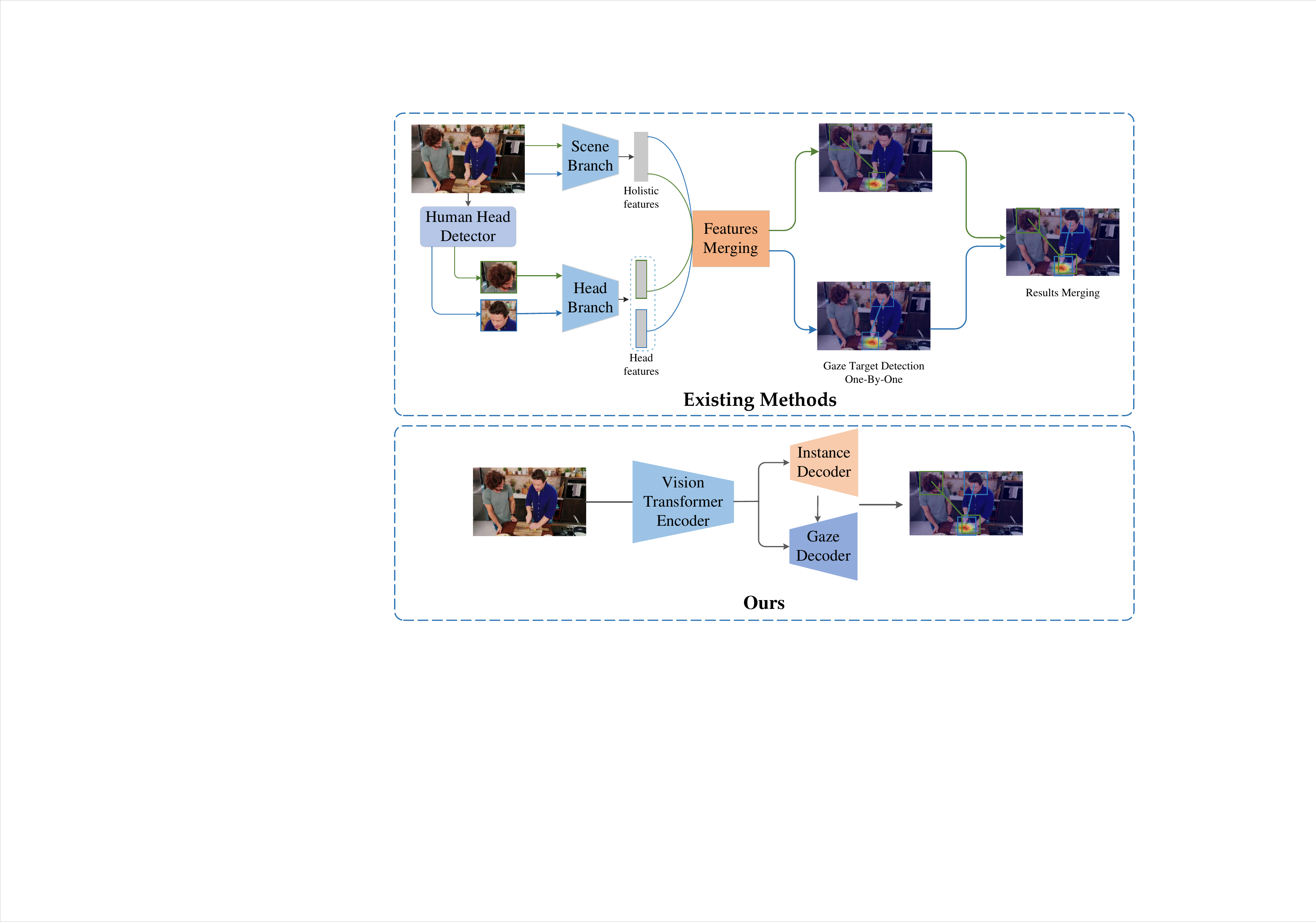}
        \captionsetup{size=small}
		\caption{{\bfseries Existing methods vs. our model}. Existing methods conventionally execute a multi-stage framework, yielding less competitive performance in model efficiency and limiting model precision due to error accumulation. In comparison, our method dispenses with any additional modules, leading to the first unified and single-stage pipeline for gaze-following detection.} 
		\label{first_img}
	\end{figure}
 
   Despite varying in targets, current gaze following detection systems invariably share a similar multi-stage framework. As shown in Fig.~\ref{first_img}, taking both a scene image and a given human head crop as inputs, recent GL-D approaches conventionally adopt a scene branch to extract holistic cues and another parallel head branch to capture head pose features. These holistic and head features are subsequently agglomerated to predict the gaze location of the given human, which is followed by a post-processing module to merge all humans' gazes in an identical scene onto one canvas. On this basis, GO-D methods further leverage an additional off-the-shelf object detector to recognize the instances being staring at. These methods are highly instructive since they demonstrate the ability of machines to estimate human gaze targets directly from images or videos,  without the necessity of any specialized sensors, such as monitor-based or wearable eye trackers.
    
    Nevertheless, these complex pipelines work slowly and show inferior precision due to several major limitations: (\emph{I}) Current methods invariably take as inputs a human head crop and a scene image, where the former is manually labeled by humans. Therefore, in real-world applications, an additional human head detector is essential to detect human head crops firstly, which enforces a two-stage framework that struggles with head crops extraction, yielding less competitive performance in model efficiency. Moreover, such a two-stage and cascaded pipeline potentially amplifies the impact of false detection by human head detector, \emph{i.e.}, a minor head detection error can fail the subsequent gaze following reasoning (Sec.~\ref{sec:pd}). (\emph{II}) Taking a given head crop as input, existing methods are restricted to detect gaze following sequentially, \emph{i.e.}, only one person's gaze target can be predicted at a time. Such limitation further degrades model efficiency when handling with multiple humans in an identical scene, as the detection process is forced to be repeatedly performed even if the holistic scene features remain unchanged (Sec.~\ref{sec:pd}). (\emph{III}) Existing methods extract holistic scene contexts and head pose features separately, lacking contextual relational reasoning for interactions between them (Sec.~\ref{sec:md}). (\emph{IV}) There is no an unified paradigm that enables to detect gaze locations and gaze objects simultaneously in an end-to-end manner, \emph{i.e.}, without using any additional components, including but not limited to the additional object detectors.

    To alleviate the above limitations, we reframe gaze following detection as a set-based prediction problem, straight from image pixels to gaze point coordinates and gaze objects, leading to a fast and accurate detection system. Therefore, we propose \textbf{GTR}, a novel \underline{\textbf{G}}aze following detection \underline{\textbf{TR}}ansformer that unites human gaze location detection and gaze object detection in an fully end-to-end framework for the first time. The superiority of our method is listed as follows:

    First, GTR is extremely fast with two novel designs: 1) We abandon the practice of using human head crops as inputs and instead leverage them as the prediction targets of the GTR. Not only does it eliminate the utilization of an extra head detector, but it also allows GTR to flexibly predict the gaze followings of humans with an arbitrary number at once. Concretely, GTR's outputs can be formulated as $N$ human-gaze-following (HGF) triplets in the format of $<$ \texttt{head location}, \texttt{gaze location}, \texttt{gaze object} $>$, where $N$ counts the number of humans in a given image.  2) GTR dispenses with all offline components, such as post-processing module and  the additional object detector, and is designed as a visual encoder followed by a two-branch decoder, which is organized as a single-stage pipeline and can be trained in a completely end-to-end manner. Hence, taking as input an image containing 6 people, our GTR is more than nine times faster than prior methods, and the relative FPS gap is more evident as the number of humans in an image grows.

    Second, GTR shows significant precision, especially in practical scenarios  with various distortions. With Transformer as key component, GTR outperforms previous methods with an excellent capability to capture long-range gaze behaviours by enhancing the contextual relational reasoning. Specifically, GTR starts with performing a visual encoder to extract global memory features, followed by a two-branch decoder (\emph{i.e.}, human decoder and gaze following decoder), where the former recognizes all humans in an image and the latter infers their gaze locations and objects. Meanwhile, a guided mechanism is developed to differentiate different human-gaze-object pairs by 
    dynamically agglomerating the human query and the global scene context, which enables an iterative and hierarchical interaction between human decoder and gaze following decoder. Finally, the HGF triplets are predicted using several multi-layer perceptrons in parallel. Besides, we devise a new objective function, which unties the GL-D and GO-D in an unified framework and allows the model to be trained in an end-to-end manner. Extensive experimental results show that GTR outperforms existing methods by significant margins. Specifically, GTR realizes a \textbf{12.1} mAP gain ($\mathbf{25.1}\%$) on GazeFollowing~\cite{recasens} and a \textbf{18.2} mAP gain ($\mathbf{43.3\%}$) on VideoAttentionTarget~\cite{chong} for GL-D, as well as  a \textbf{19.0} mAP improvement ($\mathbf{45.2\%}$) on GOO-Real~\cite{tomas2021goo} for GO-D.

    The remainder of this paper is organized as follows. Section~\ref{sec:re} reviews the prior works related to ours. Section~\ref{sec:met} explores the solution of uniting GL-D and GO-D in a single-stage pipeline. Section~\ref{sec:ex} studies the efficiency and effectiveness of our proposed method. Section~\ref{sec:dis} discusses the failed cases of our method and the potential solutions. Section~\ref{sec:con} gives the concluding remarks.

    \section{Related Work}
    \label{sec:re}

    \noindent
    \textbf{Eye tracking} was widely explored in psychological research~\cite{holmqvist2011eye, delvigne2021phydaa, zhu, kr, zhu2017, zhang} and saliency prediction tasks~\cite{duan2022saliency, paul2018spatial,ge2022tcnet, hou2007saliency, yan2013hierarchical, leifman, judd}, which aims to predict the eye fixations of an observer looking at a given picture, with the observer outside the picture. Specifically, eye tracking systems were typically built on sophisticated monitor-based or wearable eye tracking devices. As these devices require delicate pre-commissioning, the observer is expected to maintain in a stationary position with a well-calibrated posture during experiments.  In contrast, the goal of gaze following detection is to estimate what is being looked at by a person inside a picture or video~\cite{recasens}, \emph{i.e.}, detecting human attentions in an unconstrained manner. Therefore, observers are allowed to act freely in an arbitrary scenario in gaze following detection systems and we observe them in a third-person perspective. With this, gaze following detection is a more practical vision task that can be use in many daily scenarios, such as understanding human-human interactions and human-object interactions.

    \vspace{1.25mm}
    \noindent
    \textbf{Gaze following detection} was first proposed in~\cite{recasens}, which built GazeFollowing, a large-scale dataset annotated with the location of where people in images are looking. On this basis, Chong \emph{et al.}~\cite{chong2018} further solved the problem of out-of-frame gaze targets prediction (\emph{i.e.}, the gaze location of someone is not in the given image) by predicting saliency map and learning gaze angle simultaneously. Meanwhile, the precision of image-based gaze target prediction was continually improved by leveraging other different auxiliary information, such as body pose~\cite{guan2020}, sight lines~\cite{zhao2019, lian2018}, depth~~\cite{fang2021dual} and etc. In addition to detecting gaze following in image, Chong \emph{et al.}~\cite{chong} proposed a new framework to understanding human gaze following from videos, which models the dynamic interaction between the scene and head features using attention mechanism and infers time-varying attention targets. They also built a new annotated dataset named VideoAttentionTarget, which contains complex and dynamic patterns of real-world gaze behavior. There are also other related works, including but not limited to~\cite{chen2021gaze, judd, leifman, zhu, kr, hu2022gaze}.

    More recently, ~\cite{tomas2021goo} presented a new task, gaze object detection, where one must infer the bounding box of the object gazed at by the target person. It also propose a new Gaze On Objects (GOO) dataset that is composed of a large set of synthetic images (GOO-Synth) supplemented by a smaller subset of real images (GOO-Real) of people looking at objects in a retail environment.  On this basis, Wang \emph{et al.}~\cite{wang2022gatector} propose a new method GaTector for gaze object detection, which leveraged an additional object detector (YOLOV4~\cite{bochkovskiy2020yolov4}) to recognize target objects.

    Despite being ingenious, these methods invariably took manually annotated human heads as inputs, which enforces a multi-stage pipeline that inevitably compromises their efficiency. Besides, a unified framework to jointly detect human gaze location and human gaze object in an end-to-end manner awaits further exploration.

    \vspace{1.25mm}
    \noindent
    \textbf{Transformer} was originated from the domain of natural language processing~\cite{attention} , which has recently started to revolutionize the computer vision community. In ViT~\cite{dosovitskiy2020image}, the first vision Transformer, an image is first divided into $16 \times 16$ local patches.  Along with an additional class token, these local patches are fed into a Transformer encoder, which is organized as several identical self-attention layers, allowing the class token to capture long-range contextual information. Finally, taking the class token as input, a MLP is employed to predict the possible category. Besides, Transformer has also shown great potential in many other vision tasks, including but not limited to~\cite{zheng2021rethinking, strudel2021segmenter,chen2021pre,tu2022iwin, tuvideo,deformableDETR, tu2022end, tu2023agglomerative}. These methods share a similar DETR~\cite{carion2020end}-like framework. Specifically, in a DETR-like framework, a ResNet~\cite{he2016deep} backbone is first used to capture high-level visual features. Then, along with additional position embeddings, these visual features are fed into a Transformer-encoder to capture global memory features. After that, a set of learnable positional embeddings, named object queries, is responsible for interacting with the global memory features and aggregating information through several interleaved cross-attention and self-attention modules. Eventually, these object queries, decoded by a feed-forward Network (FFN), directly correspond to the final predictions. However, although Transformer has demonstrated great capability in many vision tasks, there is no Transformer-based method for gaze following detection. To the best of our knowledge, we are the first to introduce the Transformer into the domain of gaze following detection.

    \section{Methods}
    \label{sec:met}
    \subsection{Problem Reformulation}
    Given an image $\mathbf{x} \in \mathbb{R}^{3 \times H \times W}$ that contains one or more humans as the input, we redefine gaze following detection as a problem of locating of all humans at one time, as well as predicting of their gaze locations and objects. For convenience, the location of $i$-th human $\mathbf{l}_{\text{h}}^i$ is denoted as his/her head location since it is available in existing datasets. Concretely, we represent a human's head location as a rectangular bounding box $([x_{\text{tl}}, y_{\text{tl}}], [x_{\text{br}}, y_{\text{br}} ])$, where the subscripts \texttt{tl} and \texttt{br} indicate top-left and bottom-right, respectively. In addition, the gaze location $\mathbf{g}_{\text{l}}^i$ of the $i$-th human is represented as a Gaussian heatmap, of which the gaze point is located in the center. Meanwhile, we denote the gaze object of the $i$-th human as $\mathbf{g}_{\text{o}}^i$, which consists of a bounding box and an object label. In this way, our new problem can be formulated as maximizing a joint \emph{posteriori} of the output triplets $\langle \mathbf{l}_{\text{h}}, \mathbf{g}_{\text{l}}, \mathbf{g}_{\text{o}} \rangle$ on the given image $\mathbf{x}$:
    \begin{equation}
        \label{equ:1}
       \mathcal{T}^* \doteq \underset{\mathcal{T}}{\max}\,\prod_{i=1}^{N}p(\langle \mathbf{l}_{\text{h}}^i, \mathbf{g}_{\text{l}}^i, \mathbf{g}_{\text{o}}^i \rangle\mid\mathbf{x}),
    \end{equation}
where $N$ can be an arbitrary constant that counts the number of the humans in $\mathbf{x}$ and  $\mathcal{T}^*$ refers to the optimal model.

It is worth to emphasize that our new formulation is quite different from the traditional gaze following detection problem~\cite{recasens, chong2018,fang2021dual,chen2021gaze, judd, leifman, zhu, kr, hu2022gaze,chong}, which demands both scene image $\mathbf{x}$ and a given human head crop $\mathbf{x}_{\text{h}}^i$ as inputs. As aforementioned, it can only predict one individual's gaze location or object $\mathbf{l}_{\text{g}}^i/\mathbf{o}_{\text{g}}^i$ at a time, which is formulated as:
\begin{equation}
    \label{equ:2}
    \mathcal{T}_{cls}^* \doteq \underset{\mathcal{T}_{cls}}{\max}\,p(\mathbf{l}_{\text{g}}^i/\mathbf{o}_{\text{g}}^i \mid\mathbf{x},\mathbf{x}_{\text{h}}^i),
\end{equation}
where $i$ ranges from 1 to $N$. Built on this formulation, current methods handle a same image as $N$ different cases if there are $N$ different humans in the image. In comparison, our new formulation introduces a more ambitious problem, but it renders the task of gaze following detection more pertinent to practical demands, enhancing its applicability in real-world scenarios.

\subsection{Network Architecture}
In this subsection, we aim to provide the first paradigm that enables gaze location and gaze object to be detected in an unified and end-to-end manner, \emph{i.e.}, our novel GTR.

\begin{figure*}[!t]
		\centering
		\includegraphics[width=\linewidth]{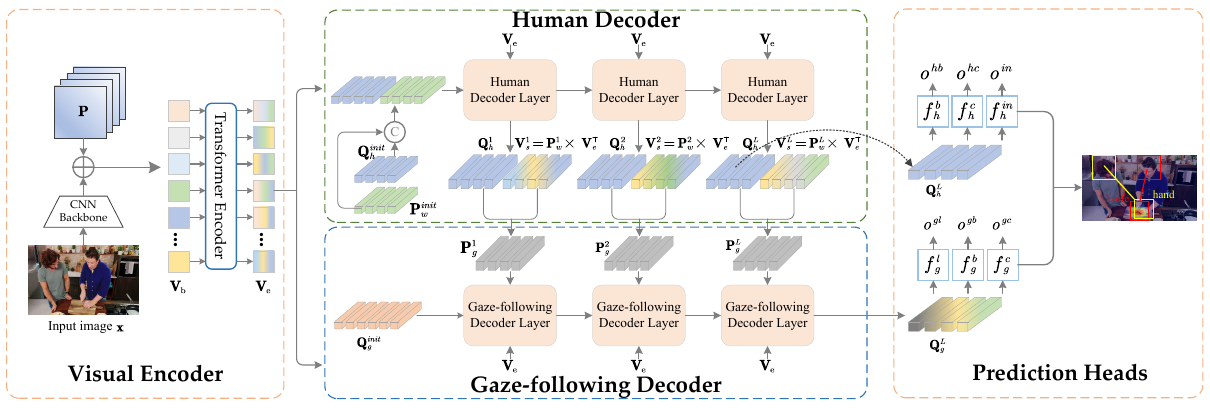}
        \captionsetup{size=small}
		\caption{{\bfseries Pipeline of our GTR}. The GTR is organized as a visual encoder equipped with two-branch decoders architecture. Given an image, the visual encoder is first applied to extract the global visual features. Then, these global visual features are fed into the subsequent two branches, \emph{i.e.}, human decoder and gaze following decoder, to localize human heads and reason their gaze followings through learnable queries, respectively. Meanwhile, we diverse a weight Guided Embedding (w-GE) to enable an iterative and hierarchical connection between these two decoders, leading to a dynamical interaction between human head features and holistic scene context. Finally, the human head crops and gaze following instances are predicted using several prediction heads in parallel.} 
		\label{whole_view}
  \vspace{-1em}
	\end{figure*}

 As illustrated in Fig.~\ref{whole_view}, GTR is organized as a visual encoder followed by two-branch decoders architecture. In visual encoder, we first adopt a CNN-based backbone to extract higher-level visual features $\mathbf{V}_b$, which are subsequently enhanced by a Transformer encoder to capture sequenced global memory features $\mathbf{V}_e$. Then, we perform two-branch decoders, \emph{i.e.}, human decoder and gaze-following decoder, to detect HGF triplets. Taking $\mathbf{V}_e$ as inputs, the human decoder detects all humans through the human query set $\mathbf{Q}_h$. Meanwhile, we design a weight Guided Embedding (w-GE) $\mathbf{P}_w$ to model the relations of all human queries with different regions in the scene. In the gaze-following decoder, we first generate gaze-following queries $\{\mathbf{Q}_g^l\}_{l=1}^{L}$ for each gaze-following decoder layer by dynamically agglomerating human queries with global context based on the w-GE \{$\mathbf{P}_w^l\}_{l=1}^{L}$ from the corresponding human decoder layer, where L denotes the number of layers. Namely, the gaze-following decoder interacts with human decoder iteratively and hierarchically, which enables a hierarchical and dynamical information  propagation from human decoder to gaze following decoder, allowing the latter to comprehend HGF triplets under the guidance of human features and holistic context. Finally, the HGF prediction results are generated by several multi-layer preceptions.

\vspace{1.25mm}
\noindent
\textbf{Visual encoder.} We follow the standard query-based Transformer detector~\cite{carion2020end} to adopt a CNN equipped with a vision Transformer encoder~\cite{dosovitskiy2020image} architecture as the visual encoder. Concretely, given an input image $\mathbf{x} \in \mathbb{R}^{3 \times H \times W}$ containing an arbitrary number of humans, we first perform a ResNet~\cite{he2016} as backbone to extract a downsampled feature map $\mathbf{V}_{\text{CNN}} \in \mathbb{R}^{2048 \times \frac{H}{32} \times \frac{W}{32}}$, which is subsequently fed into a projection convolution layer with a kernel size of $1 \times 1$ to reduce the dimension from 2048 to 256, leading to a new feature map $\mathbf{V}_{\text{CNN}}' \in \mathbb{R}^{256 \times \frac{H}{32} \times \frac{W}{32}}$. Following that, a flatten operator is used to collapse the spatial dimension into one, leading to sequenced features $\mathbf{V}_{\text{b}} \in \mathbb{R}^{256 \times \frac{HW}{32 \cdot 32}}$. Finally, a Transformer encoder containing six identical layers is employed to enhance the $\mathbf{V}_{\text{b}}$ with richer contextual information, where each layer is built upon a multi-head self-attention (MSA) module and a feed-forward network (FFN). Meanwhile, a cosine positional encoding $\mathbf{P} \in \mathbb{R}^{256 \times \frac{HW}{32 \cdot 32}}$ is additionally fed into the encoder to supplement the positional information as the Transformer architecture is permutation invariant. After the CNN-Transformer combined visual encoder, we eventually extract sequenced and global visual features $\mathbf{V}_e \in \mathbb{R}^{256 \times \frac{HW}{32 \cdot 32}}$ from the input image $\mathbf{x}$ for the following tasks.

\vspace{1.25mm}
\noindent
\textbf{Human decoder.} Our human decoder transforms $N_h^q$ human queries by stacking $L_h$ decoder layers consisting of self-attention and cross-attention mechanisms. Simultaneously, it is also expected to learn a set of weight Guided Embeddings (w-GE) for the subsequent holistic context agglomeration. As Fig~\ref{whole_view} shown, the $l$-th human decoder layer takes three parameters as inputs: 1) Global visual features $\mathbf{V}_e$ from the visual encoder, which aims at providing holistic context. 2) Human queries $\mathbf{Q}_h^l \in \mathbb{R}^{N_h^q \times C_q}$, a set of learnable parameters applied to highlight different human instances by iterative learning. Besides, to make these human queries distinct, we additionally add a set of learnable embedding $\mathbf{P}_h^l \in \mathbb{R}^{N_h^q \times C_q}$ to them, serving for extracting the interrelations between all humans instances. 3) Weight Guided Embedding (w-GE) $
\mathbf{P}_w^l \in \mathbb{R}^{N_h^q \times C_q}$.  After CNN backbone, an image is downsampled by $32 \times 32$, leading to total $\frac{HW}{32\cdot32}$ image patches with resolution of $32 \times 32$ (each pixel of the feature map from the last layer of ResNet has a receptive field of $32\times 32$). Therefore, the $n$-th vector $\mathbf{p}_w^{(l,n)} \in \mathbb{R}^{1\times C_q}$ in $\mathbf{P}_w^l$ aims to learn the relations between the $n$-th human instance with each patch, \emph{i.e.}, a local region, in the global scene. Here, $C_q = \frac{HW}{32\cdot32}$, \emph{i.e.}, the number of image patches. Formally, the human decoder is organized as:
\setlength{\abovedisplayskip}{9pt}
\setlength{\belowdisplayskip}{9pt}
\begin{align}
    \setlength{\abovedisplayskip}{10pt}
    \setlength{\belowdisplayskip}{10pt}
    [\hat{\mathbf{Q}}_h^l\;;\; \hat{\mathbf{P}}_w^l] &= \text{MSA}([\Tilde{\mathbf{Q}}_h^l;\; \mathbf{P}_w^l]),\\[1.5mm]
    [\hat{\mathbf{Q}}_h^l\;;\; \hat{\mathbf{P}}_w^l] &= \text{MCA}([\hat{\mathbf{Q}}_h^l;\; \hat{\mathbf{P}}_w^l], \mathbf{V}_e),  \\[1.5mm]
    [\hat{\mathbf{Q}}_h^l\;;\; \hat{\mathbf{P}}_w^l] &= \text{FFN}([\hat{\mathbf{Q}}_h^l; \;\hat{\mathbf{P}}_w^l]), \label{equ:human}
\end{align}
where \texttt{MSA($\cdot$)} and \texttt{MCA($\cdot, \cdot$)} refer to multi-head self-attention and multi-head cross-attention respectively. [;] denotes concatenation operator. $\Tilde{\mathbf{Q}}_h^{init}$ is initialized as the sum of $\mathbf{Q}_h^l$ and $\mathbf{P}_h^l$. Concretely, we initialize $\mathbf{Q}_h^{init}$ as zeros and $\mathbf{P}_h^{init}$, $\mathbf{P}_w^{init}$ as normal gaussian embeddings for the first layer, while the other layers take the outputs of previous layer as inputs: 
\begin{gather}
    \Tilde{\mathbf{Q}}_h^1 = \Tilde{\mathbf{Q}}_h^{init} =\underbrace{\text{Zeros}(N_h^q, C_q)}_{\mathbf{Q}_h^{init}} +  \underbrace{\text{Embedding}(N_h^q, C_q)}_{\mathbf{P}_h^{init}}, \\[1.5mm]
    \mathbf{P}_w^1 = \mathbf{P}_w^{init}= \text{Embedding}(N_h^q, C_q),\\[1.5mm]
    [\Tilde{\mathbf{Q}}_h^l\;;\;\mathbf{P}_w^l]=[ \hat{\mathbf{Q}}_h^{l-1}\;;\;\hat{\mathbf{P}}_w^{l-1}] \;\;\;\;\;\;\; if \;\;\;\; l>1.
\end{gather}

 Note that we omit the modules of residual connection and normlization in the above equations for convenience. Besides, in Fig~\ref{whole_view}, we omit the $\mathbf{P}_h^{init}$ for conciseness.

\vspace{1.25mm}
\noindent
\textbf{Gaze-following decoder.} The goal of the gaze-following decoder is to predict where and what of each human is looking at, which demands of sufficient interaction between human features and global scene contexts. Specifically, the gaze-following decoder shares a similar architecture with human decoder, which is organized as a hierarchy of $L_g$  Transformer decoder layers. By design, $L_g = L_h=L$, so we still use $l$ to indicate the index of different layers. Then, for the $l$-th gaze following decoder layer, it also takes three elements as inputs: 1) The global visual features $\mathbf{V}_e$. 2) Gaze following queries $\mathbf{Q}_g^l$, which are also learnable parameters that initialized as zeros. 3) Guided agglomeration $\mathbf{P}_g^l$ of human features and scene contexts. Concretely, we first agglomerate holistic features dynamically based on the learned inner-relations between different human instances and the whole scene via
\begin{equation}
    \mathbf{V}_s^l = \mathbf{P}_w^l \times \mathbf{V}_e^\top,
\end{equation}
where $\mathbf{P}_w^l$ is the updated w-GE in the $l$-th human decoder layer, which attends to the potential salient regions. Namely,  $\mathbf{V}_s^l$ is a selective agglomeration of global context that highlighted by human decoder. Subsequently, the $\mathbf{P}_g^l$ can be obtained as:
\begin{equation}
    \mathbf{P}_g^l = W_g [\hat{\mathbf{Q}}_h^l \;;\; \mathbf{V}_s^l].
\end{equation}

  Specifically, $\hat{\mathbf{Q}}_h^l$ are the intermediate human features extracted by the $l$-th human decoder layer (Equation (\ref{equ:human})) and $W_g$ are learnable weights of a linear layer. $\{\mathbf{P}_g^l\}_{l=1}^{L}$ for different gaze following decoder layers contain different levels of human features and different levels of holistic features, enabling a hierarchical and dynamical interaction of human features and scene contexts for the gaze following decoder. Finally, the gaze following decoder can be calculated as:
 \begin{gather}
     \hat{\mathbf{Q}}_g^l = \text{FFN}(\text{MCA}(\text{MSA}(\mathbf{Q}_g^l+\mathbf{P}_g^l),\mathbf{V}_e)),\\[1.5mm]
     \mathbf{Q}_g^1=\mathbf{Q}_g^{init}=\text{Zeros}(N_g^q, C_q),\\[1.5mm]
     \mathbf{Q}_g^l=\hat{\mathbf{Q}}_g^{l-1} \;\;\;\;\;if\;\;\; l>1.
 \end{gather}
Intuitively, $\mathbf{P}_g^l$ are explicit position encodings, which contain the features of different humans and the relevant scene contexts, guiding the gaze following queries $\mathbf{Q}_g^l$ to attend to different locations and objects for different humans.

\vspace{1.25mm}
 \noindent
 \textbf{Prediction heads.} We define each ground-truth HGF instance by the following five vectors: a human-head-bounding-box vector normalized by the corresponding image size $\mathbf{t}^{hb} \in [0,1]^4$; a watch-in-out (whether the gaze target is located inside the scene image or not) binary one-hot vector $\mathbf{t}^{io} \in \{0, 1\}^2$; a normalized gaze-object-bounding-box vector $\mathbf{t}^{gb} \in [0,1]^4$; a gaze-object-category one-hot vector $\mathbf{t}^{gc} \in \{0, 1\}^{N_o + 1}$; and  a gaze location headmap vector $\mathbf{t}^{gl} \in [0,1]^{H_o \times W_o}$, where $H_o \times W_o$ denotes the spatial resolution of the output gaze heatmap. $N_o$ is the number of object categories and the last element (the $(N_o+1)$-th element) in $\mathbf{t}^{gc}$ indicates \emph{no-object}, \emph{i.e.}, someone is looking at background.

 On this basis, we perform several multi-layer perception (MLP) on each human query and gaze-following query to decode the final HGF instance. Concretely, taking as inputs the human queries extracted by the last human decoder layer $\mathbf{Q}_h^L$, we perform an one-layer MLP branch $f_{io}$ and a three-layer MLP branch $f_h^b$ to predict the watch-in-out confidence $\mathbf{o}^{io}$ and human head bounding box $\mathbf{o}^{hb}$, respectively. Note that we additionally adopt an one-layer MLP branch $f_h^c$ to estimate the human confidence $\mathbf{o}^{hc}$, \emph{i.e.}, whether the predicted human-head-bounding-box is a true human head or not.
Synchronously, we feed the gaze following queries captured by the last gaze following decoder layer $\mathbf{Q}_g^L$ to a five-layer MLP branch $f_g^l$, a one-layer MLP branch $f_{g}^c$ and a three-layer MLP $f_g^b$ to predict human gaze location $\mathbf{o}^{gl}$, human gaze-object probability $\mathbf{o}^{gc}$ and human-gaze-object bounding box $\mathbf{o}^{gb}$, respectively. We use a softmax function for both $f_h^c,\; f_{io}, \;f_{g}^c$ and a sigmoid function for $f_h^b, \; f_g^b,\; f_g^l$.

 \subsection{Loss Calculation}
\label{loss}
 The loss calculation is composed of two stages: the bipartite matching between ground-truths and model predictions, which is followed by the loss computation for the matched pairs.

\vspace{1.25mm}
 \noindent
 \textbf{Bipartite matching.} We following the training procedure of DETR~\cite{carion2020end} to use the Hungarian algorithm~\cite{kuhn1955hungarian} for the bipartite matching, which is designed to obviate the process of suppressing over-detection. First, we pad the ground-truth of HGF instances with $\phi$ (no instance) to make the size of ground-truths set becomes $N_q$.

 As illustrated in Fig~\ref{whole_view}, our GTR outputs a fixed-size set of $N_q$ HGF instance predictions, and we denote them as $\mathbf{O}=\{\mathbf{o}_j\}_{j=1}^{N_q}$, in which the $j$-th prediction $\mathbf{o}_j = \{\mathbf{o}_j^{hb}, \: \mathbf{o}_j^{hc}, \: \mathbf{o}_j^{io}, \: \mathbf{o}_j^{gl}, \: \mathbf{o}_j^{gb}, \: \mathbf{o}_j^{gc}\}$. Meanwhile, we use $\mathbf{T} = \{\mathbf{t}_k\}_{k=1}^{N_{gt}} \bigcup \{\mathbf{t}_n=\phi\}_{n=N_{gt}+1}^{N_q}$ to represent the padded ground-truths and $\mathbf{t}_k = \{\mathbf{t}_k^{hb}, \: \mathbf{t}_k^{io}, \: \mathbf{t}_k^{gl}, \: \mathbf{t}_k^{gb}, \: \mathbf{t}_k^{gc}\}$ (human confidence has no ground-truth).  $N_{gt}$ is the real number of HGF instances in a given image. By design, $N_q$ is always larger than $N_{gt}$. On this basis, the matching process can be denoted as an injective function: $\omega_{\mathbf{T} \rightarrow \mathbf{O}}$, where $\omega(i)$ is the index of predicted HGF instance assigned to the $i$-th ground-truth. We define the matching cost as:
 \begin{equation}
     \mathcal{L}_{cost} = \sum_i^{N_q}\mathcal{L}_{match}(\mathbf{t}_i, \mathbf{o}_{\omega(i)}),
 \end{equation}
 where $\mathcal{L}_{match}(\mathbf{t}_i, \mathbf{o}_{\omega(i)})$ is a matching cost between the $i$-th ground-truth and the $\omega(i)$-th prediction. Formally, $\mathcal{L}_{match}(\mathbf{t}_i, \mathbf{o}_{\omega(i)})$ consists of three types of cost, including cost for human head prediction $\mathcal{L}_h$, cost for gaze-location prediction $\mathcal{L}_{gl}$, and cost for gaze-object prediction $\mathcal{L}_{go}$, \emph{i.e.},
 \begin{equation}
    \mathcal{L}_{match}(\mathbf{t}_i, \mathbf{o}_{\omega(i)}) = \sigma_1\mathcal{L}_h + \sigma_2\mathcal{L}_{gl} + \sigma_3\mathcal{L}_{go}.
 \end{equation}
 Firstly, 
 \begin{equation}
     \mathcal{L}_h = \alpha_h^1\mathcal{L}_h^b + \alpha_h^2\mathcal{L}_h^{c},
 \end{equation}
  where $\mathcal{L}_h^b$ is human-head-bounding-box regression loss, which is a weighted sum of GIoU~\cite{giou} loss and $L_1$ loss:
  \begin{equation}
    \mathcal{L}_h^b = \beta_{hb}^1\left \| \mathbf{t}_i^{hb} - \mathbf{o}_{\omega(i)}^{hb} \right \| - \beta_{hb}^2 \text{GIoU}(\mathbf{t}_i^{hb}, \mathbf{o}_{\omega(i)}^{hb}).
  \end{equation}
 $\mathcal{L}_h^c$ indicates is-head loss, which is defined as:
  \begin{equation}
      \mathcal{L}_h^c = -\mathbf{o}_{\omega(i)}^{hc}(0),
  \end{equation}
  where the first element $\mathbf{o}^{hc}(0)$ of $\mathbf{o}^{hc}$ indicates the model confidence about the detected human instance is positive.\\
  Secondly,
   \begin{equation}
     \mathcal{L}_{gl} = \alpha_{gl}^1\mathcal{L}_g^l + \alpha_{gl}^2\mathcal{L}^{io},
 \end{equation}
 where $\mathcal{L}_g^l$ is a $L_2$ distance between predicted human gaze location and the ground-truth one:
 \begin{equation}
     \mathcal{L}_g^l = \left \| \mathbf{t}_i^{gl} - \mathbf{o}_{\omega(i)}^{gl} \right \|_2.
 \end{equation}
 $\mathcal{L}^{io}$ stands for watch-in-out loss, which is calculated as:
 \begin{equation}
     \mathcal{L}^{io} = -\mathbf{o}_{\omega(i)}^{io}(0),
 \end{equation}
where the first element $\mathbf{o}_{\omega(i)}^{io}(0)$ of $\mathbf{o}_{\omega(i)}^{io}$ similarly refers to the model confidence about the detected human gaze point is located inside the image.\\
Finally,
\begin{equation}
    \mathcal{L}_{go} = \alpha_{go}^1\mathcal{L}_g^b + \alpha_{go}^2\mathcal{L}_g^c,
\end{equation}
where $\mathcal{L}_g^b$ is gaze-object bounding box regression loss, which is also a weighted sum of GIoU loss and $L_1$ loss:
\begin{equation}
    \mathcal{L}_g^b = \beta_{gb}^1\left \| \mathbf{t}_i^{gb} - \mathbf{o}_{\omega(i)}^{gb} \right \| - \beta_{hb}^2 \text{GIoU}(\mathbf{t}_i^{gb}, \mathbf{o}_{\omega(i)}^{gb}).
\end{equation}
$\mathcal{L}_g^c$ represents object recognition cost, which is defined as:
\begin{equation}
    \mathcal{L}_g^c = -\mathbf{o}_{\omega(i)}^{gc}(k) \:\; s.t. \;\: \mathbf{t}_{i}^{gc}(k)=1,
\end{equation}
where the $k$-th element of $\mathbf{o}_{\omega(i)}^{gc}$ is the probability that the model predicts the gaze object as the ground-truth category.

We then leverage the Hungarian algorithm to determine the optimal assignment $\hat{\omega}$ among the set of all possible permutations of $N_q$ elements $\bm{\Omega}_{N_q}$. It can be formulated as:
\begin{equation}
 \label{hun}
 \hat{\omega} = \underset{\omega \in \bm{\Omega}_{N_q}}{\arg \min}\mathcal{L}_{cost}.
\end{equation}

\noindent
\textbf{Object function.} After the optimal one-to-one matching between the ground-truths and the predictions is found, the loss to be minimized in the training phase is calculated as:
\begin{align}
    \mathcal{L} ={} & \eta_1\mathcal{L}_1+\eta_2\mathcal{L}_2+\eta_3\mathcal{L}_u+\eta_4\mathcal{L}_c,\\
     \mathcal{L}_1 ={} & \frac{1}{|\bar{\Phi}|}\sum_{i=1}^{N_q}\mathbbm{1}_{\{i \notin \Phi\}} \left[\left \| \mathbf{t}_i^{hb} - \mathbf{o}_{\hat{\omega}(i)}^{hb} \right \| + \left \| \mathbf{t}_i^{gb} - \mathbf{o}_{\hat{\omega}(i)}^{gb} \right \|\right],\\
     \mathcal{L}_2 = {} &\frac{1}{|\bar{\Phi}|}\sum_{i=1}^{N_q}\mathbbm{1}_{\{i \notin \Phi\}} \left[\left \| \mathbf{t}_i^{gl} - \mathbf{o}_{\hat{\omega}(i)}^{gl}\right \|_2 \right],\\
     \nonumber \mathcal{L}_u = {} &\frac{1}{|\bar{\Phi}|}\sum_{i=1}^{N_q}\mathbbm{1}_{\{i \notin \Phi\}} \left[ 2-\text{GIoU}(\mathbf{t}_i^{hb}, \mathbf{t}_{\hat{\omega}(i)}^{hb}) \right.\\
     &{}\qquad\qquad\qquad\qquad \left.-\text{GIoU}(\mathbf{t}_i^{gb}, \mathbf{o}_{\hat{\omega}(i)}^{gb}) \right], \\\notag
    \nonumber \mathcal{L}_c= {} & \frac{1}{N_q}\sum_{1}^{N_q}\left\{ \mathbbm{1}_{\{i \notin \Phi\}}\left[ -\log\mathbf{o}_{\hat{\omega}(i)}^{gc}(k)\right] \right.\\ 
    &{} \left.+ \mathbbm{1}_{\{i \in \Phi\}}\left[ -\log\mathbf{o}_{\hat{\omega}(i)}^{gc}(N_o + 1)\right] \right\} \; s.t. \; \mathbf{t}_{i}^{gc}(k)=1,
\end{align}
where $\Phi$ is a set of ground-truth indices that correspond to $\phi$ and $\bar{\Phi}$ is its complementary set. Note that our new loss is quite different from the prior one ( typically a simple $L_2$ loss), which demonstrates a desirable effectiveness (Sec.~\ref{sec:ex}).

\section{Experiments}
\label{sec:ex}
\subsection{Datasets \& Evaluation Metrics}
	
	\noindent
	{\bfseries Datasets.} We train and test our model for GL-D on both GazeFollowing~\cite{recasens} dataset and VideoAttentionTarget~\cite{chong} dataset. Specifically, we use every single frame in VideoAttentionTarget as input during the training process, without considering the temporal information. Therefore, to avoid overfitting, for every five continuous frames from the training set of VideoAttentionTarget which have no obvious appearance differences, we randomly select one for training since they have almost the same gaze target. For testing, we still use all images in the testing set. 

    Meanwhile, we use GOO~\cite{tomas2021goo} dataset to evaluate the performance of our model for GO-D. Concretely, GOO contains annotations of bounding boxes for all gaze objects, which are labeled with 24 different categories. Additionally, GOO consists of two different subsets, \emph{i.e.}, GOO-Synth that contains 192,000 synthetic images and GOO-Real that contains 9,552 real images. 
	
	Moreover, one of objectives of our proposed model is to predict the locations of different individuals. Therefore, the head locations in existing dataset annotations are no longer used as inputs but as ground-truths. Besides, previous works predict the gaze target for different subjects in an identical scene on a case-by-case basis, which results in each image being assigned with $M$ annotations, and $M$ is the number of people in an image. Unlike that, we merge the annotations of the same image into the same format as COCO~\cite{lin2014microsoft} object detection since we aim to predict them all at one time.

 \vspace{1.1mm}
	\noindent
	{\bfseries Evaluation metric.} In this work, we have to evaluate the performance of proposed model in terms of both gaze following detection (GL-D and GO-D) and human position detection. 
	
	First, we follow the standard evaluation protocols, as in \cite{recasens,chong, tomas2021goo}, to report the results of GL-D in terms of $\textbf{AUC}$ and $\bm{L_{2}}$ distance. For AUC: the final heatmap provides the prediction confidence score which is evaluated at different thresholds in the ROC curve. The area under curve (AUC) of the ROC is reported \cite{chong}. In terms of distance: $L_2$ distance between the annotated target location and the prediction given by the pixel having the maximum value in the heatmap, with image width and height normalized to 1, is reported. Specifically, since the ground truth for GazeFollowing may be multimodal, the $L_2$ distance is the Euclidean distance between our prediction and the average of ground-truth annotations. Besides, the minimum distance between our prediction and all ground-truth annotations is also reported. In addition, the average precision (\textbf{AP}) is used to evaluate the performance for is-watching-outside prediction. On this basis, we further report the AP of object detection for the GO-D. 
	
	Then, we use the commonly used role \emph{mean average percision} (\textbf{mAP}) to examine the model performance in detecting gaze following and localizing human head simultaneously. Specifically, a HGF detection is considered as true positive if and only if it localizes the human and detects gaze following accurately (\emph{i.e.} the \emph{Interaction-over-Union} (IOU) ratio between the predicted human-head-box and ground-truth is greater than 0.5 while the $L_2$ distance for GL-D is less than 0.15 and confidence for GO-D is larger than 0.75).

 \subsection{Implementation Details}
 \vspace{1.25mm}
 \noindent
 \textbf{Model structure.} We adopt ResNet-50 and ResNet-101 as the CNN backbone of the feature extractor. Both multi-head self-attention and cross-attention has 8 heads. The number of both the human queries and the gaze following queries $N_q$ are set to 20. The number of the human decoder layers $L_h$ and the gaze following decoder layers $L_g$ are both equal to 3.

 \vspace{1.25mm}
 \noindent
 \textbf{Model Training.} We initialize the network with the parameters of DETR~\cite{carion2020end}. GTR is trained for 100 epochs using AdamW~\cite{adamw} optimizer with batch size of 16. Specifically, the initial learning rate of the backbone network is set as $10^{-5}$ while that of the others is set as $10^{-4}$, the weight decay is equal to $10^{-4}$. Both learning rates are decayed after 80 epochs. For the hyper-parameters of the model, we set $\alpha_{\{h,gl\}}^1$ and $\alpha_{\{h,gl\}}^2$ as 2 and 1 respectively, while revising them for $\alpha_{go}^1$ and $\alpha_{go}^2$. $\beta_{\{hb, gb\}}^1$ and $\beta_{\{hb, gb\}}^2$ are set to 1 and 2.5, respectively. The weights ($\sigma_1 - \sigma_3$) for different cost functions in the loss are set as 2, 1, 1, respectively, while $\eta_1 - \eta_4$ are equal to 2.5, 1, 1, 2. All experiments are conducted on 8 NVIDIA RTX 2080Ti GPUs.

 \subsection{Performance Comparison to State-of-the-Art Methods}
 \begin{table*}[htb]
  \centering
  \tablestyle{4pt}{1.2}
    \captionsetup{size=small}
  \caption{
  	\textbf{Quantitative comparisons on the GazeFollowing and VideoAttentionTarget.} 
  	We report the results of different methods  \emph{w.r.t} the ``Default" and  the ``Real" two diverse sets, respectively. Specifically, ``Default" refers to the utilization of manually annotated head locations from existing datasets, while ``real" involves the application of an additional head detector to extract head crops for real world applications.
  }
  \begin{tabular}{y{67}x{22}x{22}x{22}x{22}x{22}x{22}x{22}x{10}x{22}x{22}x{22}x{22}x{22}x{22}x{22}}
    \toprule[0.7pt]

    \multirow{3}{*}[-1em]{Method} &  
    
     \multicolumn{7}{c}{GazeFollowing~\cite{recasens}} & & 
     
     \multicolumn{7}{c}{VideoAttentionTarget~\cite{chong}} \\
     
      \cmidrule(lr){2-8}
      \cmidrule(lr){10-16}
    & \multicolumn{2}{c}{AUC$\uparrow$} & \multicolumn{2}{c}{Average Dist.$\downarrow$} & \multicolumn{2}{c}{Min Dist.$\downarrow$} & \multirow{2}{*}[-0.5em]{\textbf{mAP}$\uparrow$} & 
    & \multicolumn{2}{c}{AUC$\uparrow$} & \multicolumn{2}{c}{L2 Dist.$\downarrow$}      & \multicolumn{2}{c}{AP$\uparrow$}        & \multirow{2}{*}[-0.5em]{\textbf{mAP}$\uparrow$}  
    \\
    \cmidrule(lr){2-3}
    \cmidrule(lr){4-5}
    \cmidrule(lr){6-7}
    \cmidrule(lr){10-11}
    \cmidrule(lr){12-13}
    \cmidrule(lr){14-15}
   & Default & \textbf{Real} & Default & \textbf{Real} & Default & \textbf{Real} & &  
   & Default & \textbf{Real} & Default & \textbf{Real} & Default & \textbf{Real} & \\
    \midrule[0.5pt]
                
              Random %
              & 0.504 & 0.391 & 0.484 & 0.533 & 0.391 & 0.487 & 0.104 & 
              & 0.505 & 0.247 & 0.458 & 0.592 & 0.621 & 0.349 & 0.091
              \\
       
              Center %
              & 0.633 & 0.446 & 0.313 & 0.495 & 0.230 & 0.371 & 0.117 & 
              & ---   & ---   & ---   & ---   & ---   & ---   & ---
              \\
              
        	  Fixed bias %
        	  & ---   & ---   & ---   & ---   & ---   & ---   & --- & 
        	  & 0.728 & 0.522 & 0.326 & 0.472 & 0.624 & 0.510 & 0.130
        	  \\
        	  
        	  Judd~\cite{judd} %
        	  & 0.711 & ---   & 0.337 & ---   & 0.250 & ---   & --- & 
        	  & ---   & ---   & ---   & ---   & ---   & ---   & ---
        	  \\
        	  
        	  GazeFollow~\cite{recasens} %
        	  & 0.878 &0.804  & 0.190 & 0.233 & 0.113 & 0.124 & 0.457 & 
        	  & ---   & ---   & ---   & ---   & ---   & ---   & ---
        	  \\
        	  
        	  Chong~\cite{chong2018} %
        	  & 0.896 & 0.807 & 0.187 & 0.207 & 0.112 & 0.120 & 0.449 & 
        	  & 0.830 & 0.791 & 0.193 & 0.214 & 0.705 & 0.651 & 0.374
        	  \\
        	  Zhao~\cite{zhao2019} %
        	  & ---   &---   & 0.147  &---    & 0.082 &---    & --- & 
        	  & ---   & ---  & ---    & ---   & ---   & ---   & ---
        	  \\
        	  
        	  Lian~\cite{lian2018} %
        	  & 0.906 & 0.881 & 0.145 & 0.153 & 0.081 & 0.087 & 0.469 & 
        	  & 0.837 & 0.784 & 0.165 & 0.172 & ---   &---    & 0.392
        	  \\
        	  
        	  VideoAttention~\cite{chong} %
        	  & 0.921 & 0.902 & 0.137 & 0.142 & 0.077 & 0.082 & 0.483 & 
        	  & 0.860 & 0.812 & 0.134 & 0.146 & 0.853 & 0.849 & 0.420
        	  \\
        	  
        	  DAM~\cite{fang2021dual} %
        	  & 0.922 & ---   & 0.124 & ---   & 0.067 & ---   & --- & 
        	  & 0.905 & ---   & 0.108 & ---   & 0.896 & ---   & ---
        	  \\
        	  
        \midrule
        \rowcolor{baselinecolor} 
        	  GTR (ResNet-50)
        	  & 0.925  & \textbf{0.928}   & 0.119  & \textbf{0.114} & 0.063   & \textbf{0.057}  & \textbf{0.604} & 
        	  & 0.927  & 0.925   & 0.102  & \textbf{0.093} & 0.910   & 0.923 & 0.584
        	  \\
        \rowcolor{baselinecolor} 
        	  GTR (ResNet-101)
        	  & 0.921  & 0.925  & 0.121  & 0.116 & 0.068   & 0.064 & 0.592 & 
        	  & 0.931  & \textbf{0.936}   & 0.105  & 0.095 & 0.914   & \textbf{0.927} & \textbf{0.602}
        	  \\
        \bottomrule[0.7pt]

  \end{tabular}
\label{tab_gl}
\end{table*}

 \begin{table*}[htb]
  \centering
  \tablestyle{4pt}{1.2}
  \captionsetup{size=small}
  \caption{
  	\textbf{Quantitative comparisons on the GOO dataset.} We report the results for GL-D on the GOO-Synth subset and the results for GO-D on the GOO-Real subset, which are also based on the aforementioned ``Default" and ``Real" scenarios.
  }
  \begin{tabular}{y{67}x{22}x{22}x{22}x{22}x{22}x{22}x{22}x{10}x{22}x{22}x{22}x{22}x{22}x{22}x{22}}
    \toprule[0.7pt]

    \multirow{3}{*}[-1em]{Method} &  
    
     \multicolumn{7}{c}{GOO-Synth} & &
     
     \multicolumn{7}{c}{GOO-Real} \\
     
      \cmidrule(lr){2-8}
      \cmidrule(lr){10-16}
    & \multicolumn{2}{c}{AUC$\uparrow$} & \multicolumn{2}{c}{Average Dist.$\downarrow$} & \multicolumn{2}{c}{Ang.$\downarrow$} & \multirow{2}{*}[-0.5em]{\textbf{mAP}$\uparrow$} & 
    & \multicolumn{2}{c}{AP$\uparrow$} & \multicolumn{2}{c}{$\text{AP}_{50}$$\uparrow$}      & \multicolumn{2}{c}{$\text{AP}_{75}$$\uparrow$}        & \multirow{2}{*}[-0.5em]{\textbf{mAP}$\uparrow$}  
    \\
    \cmidrule(lr){2-3}
    \cmidrule(lr){4-5}
    \cmidrule(lr){6-7}
    \cmidrule(lr){10-11}
    \cmidrule(lr){12-13}
    \cmidrule(lr){14-15}
   & Default & \textbf{Real} & Default & \textbf{Real} & Default & \textbf{Real} & &
   & Default & \textbf{Real} & Default & \textbf{Real} & Default & \textbf{Real} & \\
    \midrule[0.2pt]
                
              Random %
              & 0.497 & --- & 0.454 & --- & 77.0 & --- & 0.076 & 
              & --- & --- & --- & --- & --- & --- & ---
              \\

        	  GazeFollow~\cite{recasens} %
        	  & 0.929 &0.832  & 0.162 & 0.317 & 33.0 & 42.6 & 0.468 & 
        	  & 35.5   & 31.6   & 58.9   & 54.3  & 26.7   & 24.1   & 0.291
        	  \\

        	  Lian~\cite{lian2018} %
        	  & 0.954 & 0.903 & 0.107 & 0.153 & 19.7 & 28.1 & 0.454 & 
        	  & 35.7 & 32.1 & 62.8 & 58.4 & 29.4   &27.2    & 0.297
        	  \\
        	  
        	  VideoAttention~\cite{chong} %
        	  & 0.952 & 0.912 & 0.075 & 0.143 & 15.1 & 24.5 & 0.489 & 
        	  & 36.8 & 32.4 & 61.5 & 56.7 & 30.8 & 27.9 & 0.314
        	  \\
                 GaTector~\cite{wang2022gatector} %
                 & 0.957 & 0.918 & 0.073 & 0.139 & 14.9 & 24.5 & 0.510 & 
        	  & 52.25 & 48.7 & 91.92 & 86.4 & 55.34 & 52.1 & 0.437\\
        	  
        \midrule
        	  \baseline{GTR (ResNet-50)}
        	  & \baseline{0.960}  & \baseline{\textbf{0.962}}   & \baseline{0.071 } & \baseline{\textbf{0.068}} & \baseline{14.5}   & \baseline{\textbf{14.2}} & \baseline{\textbf{0.597}} & \baseline{}
        	  & \baseline{58.26}  & \baseline{57.12}   & \baseline{97.31}  & \baseline{96.47} & \baseline{58.42 }  & \baseline{58.90} & \baseline{\textbf{0.627}}
        	  \\ 
        	  \baseline{GTR (ResNet-101)}
        	  & \baseline{0.954}  & \baseline{0.953}  & \baseline{0.074}  & \baseline{0.070} & \baseline{14.5 }  & \baseline{14.8} & \baseline{0.588} & \baseline{}
        	  & \baseline{58.28}  & \baseline{\textbf{58.32}}   & \baseline{96.93}  & \baseline{\textbf{96.84}} & \baseline{59.81 }  & \baseline{\textbf{59.79}} & \baseline{0.625}
        	  \\
        \bottomrule[0.7pt]

  \end{tabular}
\label{tab_go}
\end{table*}

\noindent
\textbf{Gaze location detection.} We first verify the effectiveness of our GTR in Table~\ref{tab_gl}, which compares several popular GL-D methods on GazeFollowing~\cite{recasens} and VideoAttentionTarget~\cite{chong} datasets, respectively. As aforementioned, current methods are designed to take human head crops as inputs, we thereby report the results of them on two different setting: ``Default" and "Real". For the former, we directly leverage the manually labeled head crops in dataset annotations. For the latter, we consider the practical scenario to adopt an additional off-the-shelf head detector to individually extract the head crops and feed the predicted results into existing models. In this work, we fine-tune a SSD-based~\cite{liu2016ssd} detector using the head annotations in existing dataset, which is followed~\cite{chong}, a competitive GL-D model in existing methods. As the table shown, GTR outperforms current methods on all datasets. Specifically, on Gazefollowing, GTR yields an impressive mAP gain compared to prior methods, ranging from $25\%$ to $35\%$ and leading to a new optimal result (0.604). Moreover, the results of GTR at the ``Real" setting even exceed the results of prior methods at the ``Default" setting, which are supposed to be their theoretical upper value. Note that our GTR does not require head crops as inputs, but we still report the results of GTR at the ``Default" setting by feeding both scene image and head crop into our visual encoder and invalidating the human decoder, which yet leads to an inferior results. Actually, detecting the head crops facilitates the model capability to understanding gaze following, as it allows the model to explicitly attends to head features. Additionally, we find that GTR achieves a more desirable results on VideoAttentionTarget, which can attribute to two major reasons: 1) VideoAttentionTarget is a larger dataset than GazeFollowing, which is preferred by Transformer-based methods. 2) There are more than one human in each image of VideoAttentionTarget while images in GazeFollowing contain only one person. As our model outputs a fixed volume of HGF instances, fewer human instances in an image potentially refers to a higher fault positive (FP).

\begin{figure*}[t]
    \centering
     \subfloat[
{\footnotesize \textbf{efficiency vs. human number.}}
\label{fig:eff}
]{
    \begin{minipage}[t]{0.33\linewidth}
    \centering
    \includegraphics[width=0.97\linewidth]{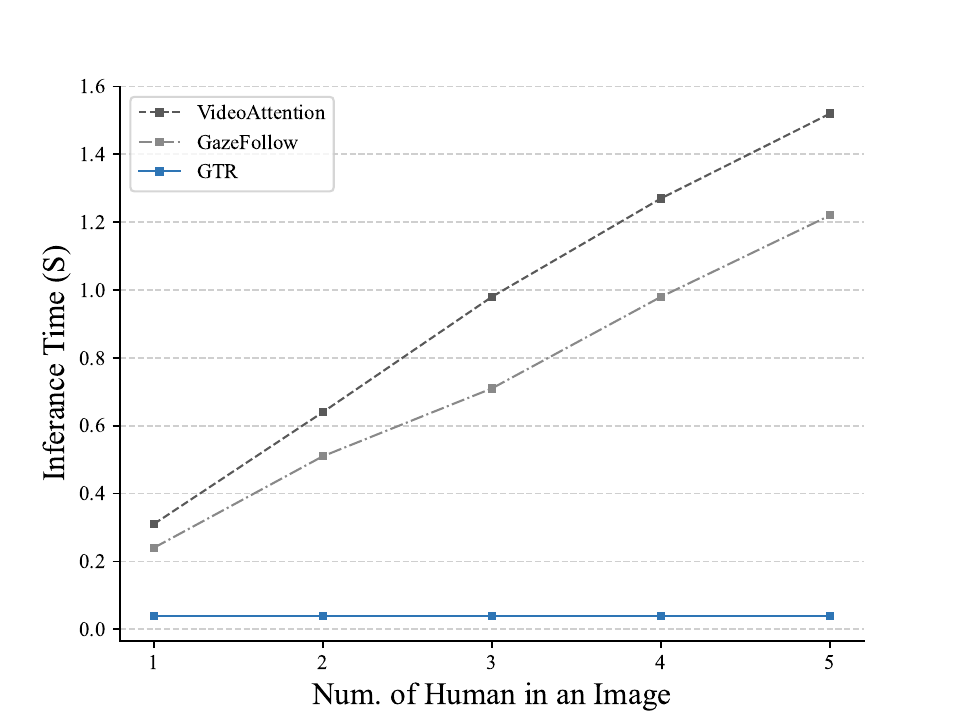}
    \end{minipage}
    }
    \subfloat[
{\footnotesize\textbf{AUC vs. distance.}}
\label{fig:dis}
]{
    \begin{minipage}[t]{0.33\linewidth}
    \centering
    \includegraphics[width=0.97\linewidth]{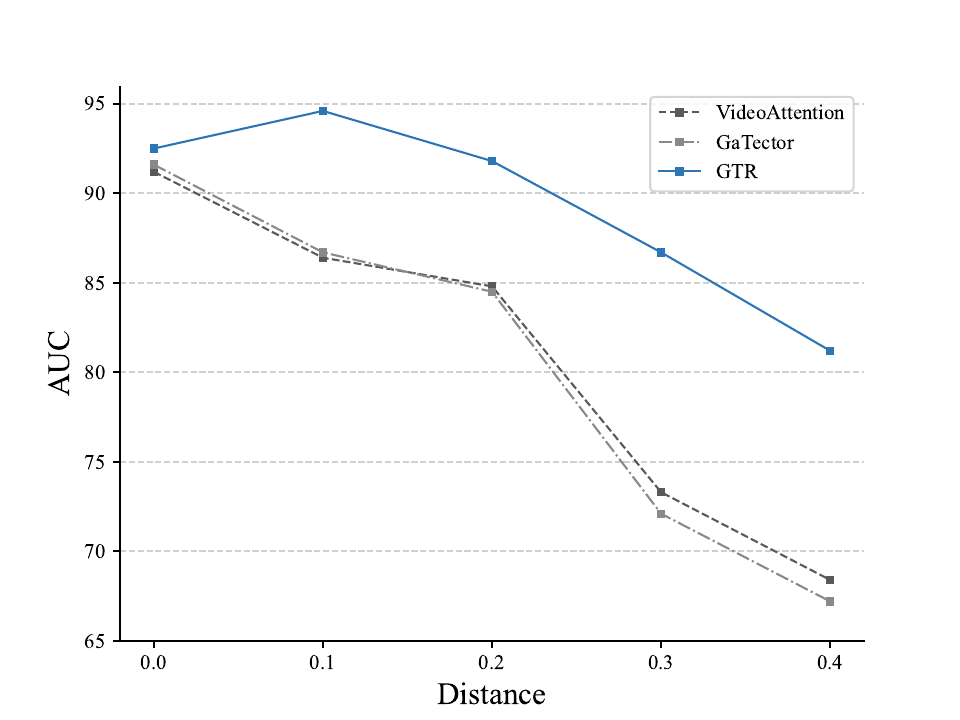}
    \end{minipage}
    }
   \subfloat[
{\footnotesize\textbf{performance vs. decoder layer number}} 
\label{fig:dec}
]{
    \begin{minipage}[t]{0.33\linewidth}
    \centering
    \includegraphics[width=0.97\linewidth]{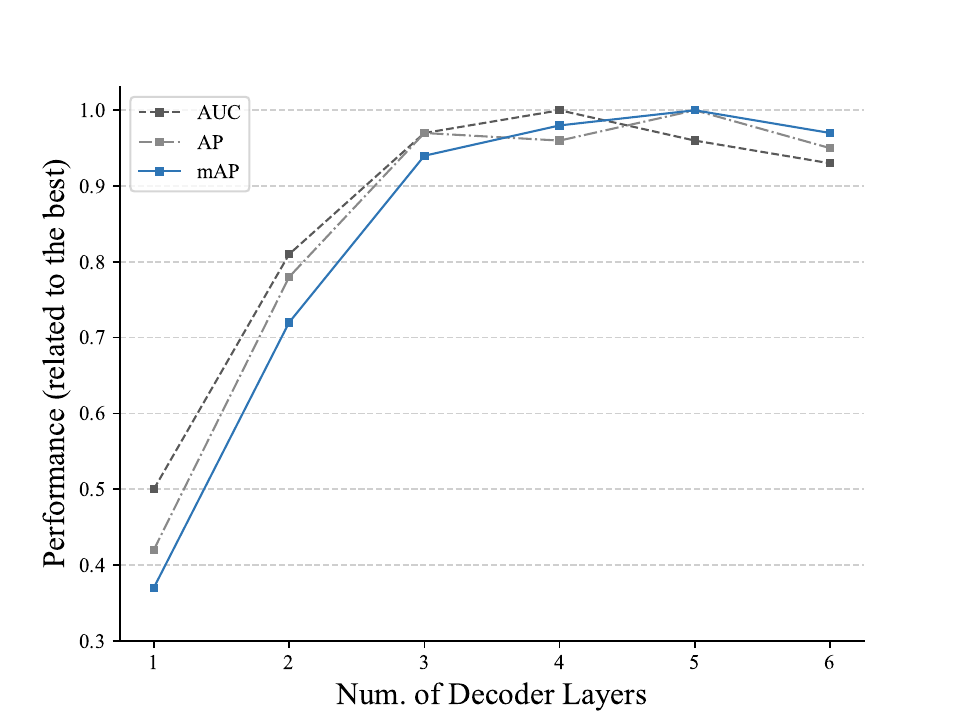}
    \end{minipage}
    }
    \label{fig:all}
    \captionsetup{size=small}
\caption{(a): Inference times of different methods with different numbers of humans in an image. GTR specializes in handling scene image containing multiple humans. (b): AUCs of different methods on different spatial distributions of HGF instances. GTR is more effective at detecting gaze followings that are more distantly distributed. (c): Performance of GTR with different numbers of decoder layers, which shows all metrics' trends as the number of decoder layers increases, \emph{i.e.}, all metrics are normalized by dividing by their maximum value.}
\end{figure*}

\begin{table*}[t]
	\centering
  \tablestyle{4pt}{1.3}
  \captionsetup{size=small}
  \caption{
  	{\bfseries Performance of using different human head detectors}. We manually degrade the original images in the testing set of VideoAttentionTarget and GOO-Real with several common distortion types. Specifically, ``FT" refers to that the training sets of those two dataset are also augmented with these distortion types and the head detectors are further fine-tuned with the head locations in the augmented dataset. ``Normal" means that no degeneration image is used.
  }
  \begin{tabular}{y{50}x{20}x{20}x{5}x{20}x{20}x{20}x{5}x{20}x{20}x{20}x{5}x{20}x{20}x{20}x{5}x{20}x{20}x{20}}
  \toprule[0.7pt]
		\multirow{2}{*}[-0.3em]{Method} & \multirow{2}{*}[-0.30em]{FT}  & \multirow{2}{*}[-0.30em]{FPS} & & \multicolumn{3}{c}{Blur} & & \multicolumn{3}{c}{Gaussian Noise} & & \multicolumn{3}{c}{Brightness} & &\multicolumn{3}{c}{Normal}\\
		\cline{5-7} 
		\cline{9-11}
		\cline{13-15}
		\cline{17-19}
		& & & & AUC & AP   & mAP  &    & AUC   & AP  & mAP    &  & AUC   & AP  & mAP    &  & AUC   & AP & mAP \\
		\hline

		  \multirow{2}{*}{SSD-based~\cite{liu2016ssd}}
		& \xmark   & 2   &   & 0.729 & 27.1  & 0.302    &  & 0.706  & 28.3 & 0.298    &  & 0.738 & 26.7  & 0.315    &  & 0.789  & 28.5 & 0.405 \\
		& \cmark   & 2   &   & 0.774 & 29.2  & 0.371   &   & 0.767  & 28.7 & 0.362   &   & 0.784 & 30.1  & 0.376   &   & 0.812  &32.4 & 0.420  \\
		\hline
		\multirow{2}{*}{FCHD~\cite{vora2018fchd}}
		& \xmark   & 3   &   & 0.757 &27.8  & 0.336   &   & 0.744 &29.2  & 0.328   &   & 0.764 &27.6  & 0.341   &   & 0.804 &30.9  & 0.409 \\
		& \cmark   & 3   &   & 0.784 &30.3  & 0.385   &   & 0.781 & 31.8  & 0.325   &   & 0.796 &30.4  & 0.392   &  &  0.818  & 33.7 & 0.424  \\
		\hline
    		\multirow{2}{*}{HeadHunter~\cite{sundararaman2021tracking}}
		& \xmark   & 1   &   & 0.782 &29.9  & 0.341   &   & 0.750 &30.6  & 0.332   &   & 0.773 & 29.8  & 0.352   &   & 0.796 & 32.1 & 0.412 \\
		& \cmark   & 1   &   & 0.787 &31.6  & 0.389   &   & 0.789 &33.9  & 0.384   &   & 0.804 &33.7  & 0.410   &   & 0.819 &35.4  & 0.424  \\
		\hline
		\rowcolor{baselinecolor}
		& \xmark   & 27  &   & 0.814 & 53.4  & 0.442   &   & 0.792 &54.1  & 0.438   &   & 0.813 &54.3  & 0.451    &    & 0.927  &57.1 & 0.584 \\
		\rowcolor{baselinecolor}
		\multirow{-2}{*}{GTR}
		 & \cmark   & \textbf{27}  &   & \textbf{0.920} & \textbf{55.6} & \textbf{0.567}   &   & \textbf{0.894} &\textbf{53.2}   & \textbf{0.513}   &  & \textbf{0.918}  &\textbf{55.8} & \textbf{0.496}   &     & ---     & ---  & ---  \\
		 	                                 
	\bottomrule[0.7pt]
		
	\end{tabular}
	\label{tab:head}
\end{table*}

\vspace{1.25mm}
\noindent
\textbf{Gaze object detection.} Table~\ref{tab_go} verifies the effectiveness of GTR for GO-D on GOO~\cite{tomas2021goo} dataset. Compared with GaTector~\cite{wang2022gatector}, GTR achieves a 9.62 AP gain ($+19.8\%$) for GO-D and 0.19 mAP gain ($+43.5\%$) for joint GO-D and human head detection on GOO-Real subset. Moreover, GTR is designed as a single-stage and end-to-end pipeline while prior methods invariably require an additional object detector, which is a fine-tuned YOLO-V4~\cite{bochkovskiy2020yolov4} by following standard protocols. Dispensing with extra detector enforces the model to attends to global contextual semantic, which is further enhanced by attention mechanism in Transformer, promising GTR an excellent performance for GO-D.

\subsection{Importance of Pairwise Detection}
\label{sec:pd}
In this work, we redefine gaze following detection as a task to jointly detect human head locations and their gaze followings. This subsection aims to validate the importance of such a pairwise detection strategy in terms of model robustness and model efficiency.

\vspace{1.25mm}
\noindent
\textbf{Model robustness}. An additional human head detector is essential for existing methods since they take both scene image and human head crop as inputs. Nevertheless, in such a cascaded pipeline, the precision of the head detector can easily compromise the performance of the gaze following detection due to potential mis-detection.  In practical applications (\emph{e.g.}, video surveillance), image degradations, such as blur, noise, brightness variation, are inevitable. Therefore, we apply several diverse head detectors to analyze model performance in different degradation conditions and report the results in Table~\ref{tab:head}. Here, we adopt VideoAttention~\cite{chong} equipped with YOLO-V4 as baseline.  As the table shown, pairwise detection not only improves the FPS by a large margin (27 vs. 2), but is also more robust with superior performance in different degradation conditions. First, using an extra head detector is a sub-optimal solution since existing models are invariably trained with head crops that are carefully and manually annotated, making they are quite sensitive to the results of the head detector. Secondly, while the influences introduced by image degradation can be partly alleviated by data augmentation in training progress, but it is hard to fine-tune the head detector in real-world applications due to the lack of manually labeled head crops. In comparison, GTR solves gaze following detection in an absolute end-to-end manner, dispensing with the extra head detector, which significantly improves both the results of GL-D (reported \emph{w.r.t} AUC, $>16.9\%$ gain) and GO-D (reported \emph{w.r.t} AP, $>60.6\%$ gain) in all degradation conditions.

\vspace{1.25mm}
\noindent
\textbf{Model efficiency}. Pairwise detection also introduces high efficiency. Using human head images as input, as existing methods do, limits the ability of detector to predict different people's gaze followings at the same time. As shown in Fig~\ref{fig:eff}, as the number of individuals in an identical scene increases, the inference time of existing methods grows extremely. The main reason is that existing methods can predict only one human's gaze target at a time,  and the detection process has to be conducted repeatedly when there are more than one human in a same image. On the contrary, our GTR is free with such limitation and its inference time is almost not inflected by the number of humans. In this paper, GTR is designed to detect up to 20 peoples gaze targets at a time. In real world applications, the model can be easily fine-tuned to increase the number of maximum detectable humans by simply changing the value of hyper-parameter $N_q$.

\subsection{Ablation Study for Model Design}
\label{sec:md}
In this subsection, we conduct extensive ablation experiments on VideoAttentionTarget and GOO-Real to validate the effectiveness of our proposed strategies for model design. All experiments are conducted using ResNet-50 as backbone.

\vspace{1.25mm}
\noindent
\textbf{Visual encoder}. Following DETR~\cite{carion2020end}, our visual encoder is designed as a CNN-based backbone followed by a Transformer encoder, whose impacts are verified in Table~\ref{tab:ve}. Comparing to feeding original image patches into Transformer, CNN-based feature extractor enables to capture higher-level semantic, leading to  a 0.176 ($\triangle \: 23.4\%$) AUC gain for GL-D and a 14.48 ($\triangle \: 34.0\%$) AP gain for GO-D. Moreover, Transformer encoder contributes a 0.243 ($\triangle \: 35.5\%$) AUC improvement and a 22.90 ($\triangle \: 66.9\%$) AP gain, which are mainly attributed to the self-attention mechanism, thanks to its excellent ability to modeling long range context. We further explore in which cases self-attention especially enables superior performance compared with the current methods. To this end, we split all HGF instances in VideoAttentionTarget test set into bins of size  0.1 on the basis of $L_1$ distance between the center of human head and her/his gaze location, where the height and width of image have been normalized. We report the results in Fig~\ref{fig:dis}, where the result of VideoAttention~\cite{chong} and GaTector~\cite{wang2022gatector} are reproted in ``Default" setting, \emph{i.e.}, using the off-the-shelf head crops as inputs. As shown in the Figure, the degraded performance indicts that the gaze following detection tends to be more difficult as the distance grows, and the relative performance gaps between the GTR and the other two detectors correspondingly become more and more evident. Particularly, our GTR shows impressive performance in distant gaze following detection while existing methods are failed. The potential reason for such result is that existing methods deal with limited receptive fields, being weak in extracting long range contextual information and being likely to be dominated by irrelevant information in the distant cases.

\begin{figure}[!t]
		\centering
		\includegraphics[width=\linewidth]{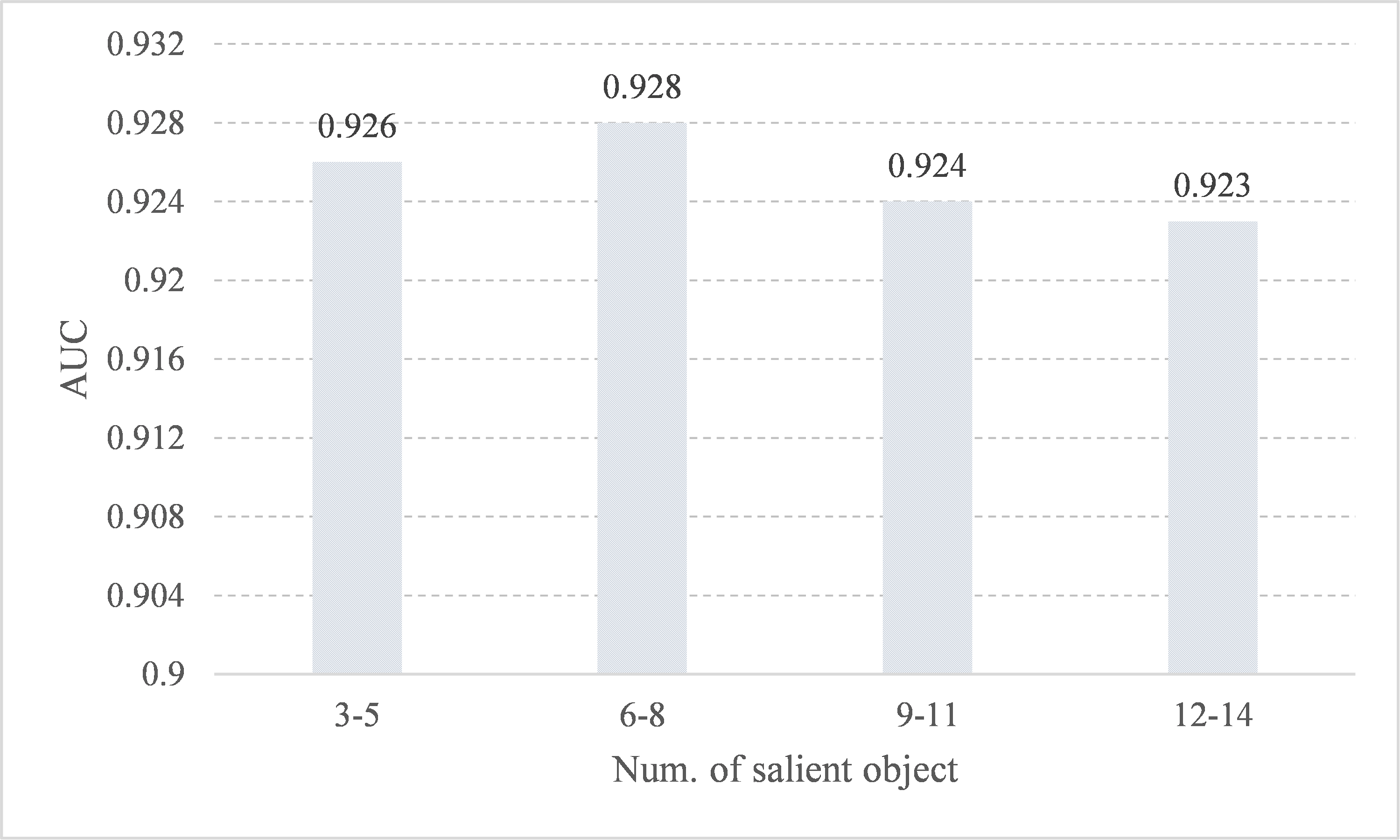}
        \captionsetup{size=small}
		\caption{ The impacts of the number of salient objects.} 
		\label{fig:nums}
  \vspace{-0.5em}
	\end{figure}

\vspace{1.25mm}
\noindent
\textbf{Two-branch decoders}. We verify the impacts of the two decoders, respectively, \emph{i.e.}, human and gaze following decoder.

\vspace{0.1mm}
\noindent
\emph{1) Human decoder}.
Table~\ref{tab:ds} verifies the effectiveness of human decoder. Comparing to using a single decoder to predict human heads and understanding their gaze following, additional human decoder introduce a $2.2\%$ AUC gain for GL-D and a $21.1\%$ AP gain for GO-D. Human head detection demands relatively focused receptive field compared to gaze following detection. Therefore, using an additional decoder can effectively reduce the potential semantic ambiguity introduced by the receptive field gaps of these two tasks.

\begin{figure*}[!t]
		\centering
		\includegraphics[width=\linewidth]{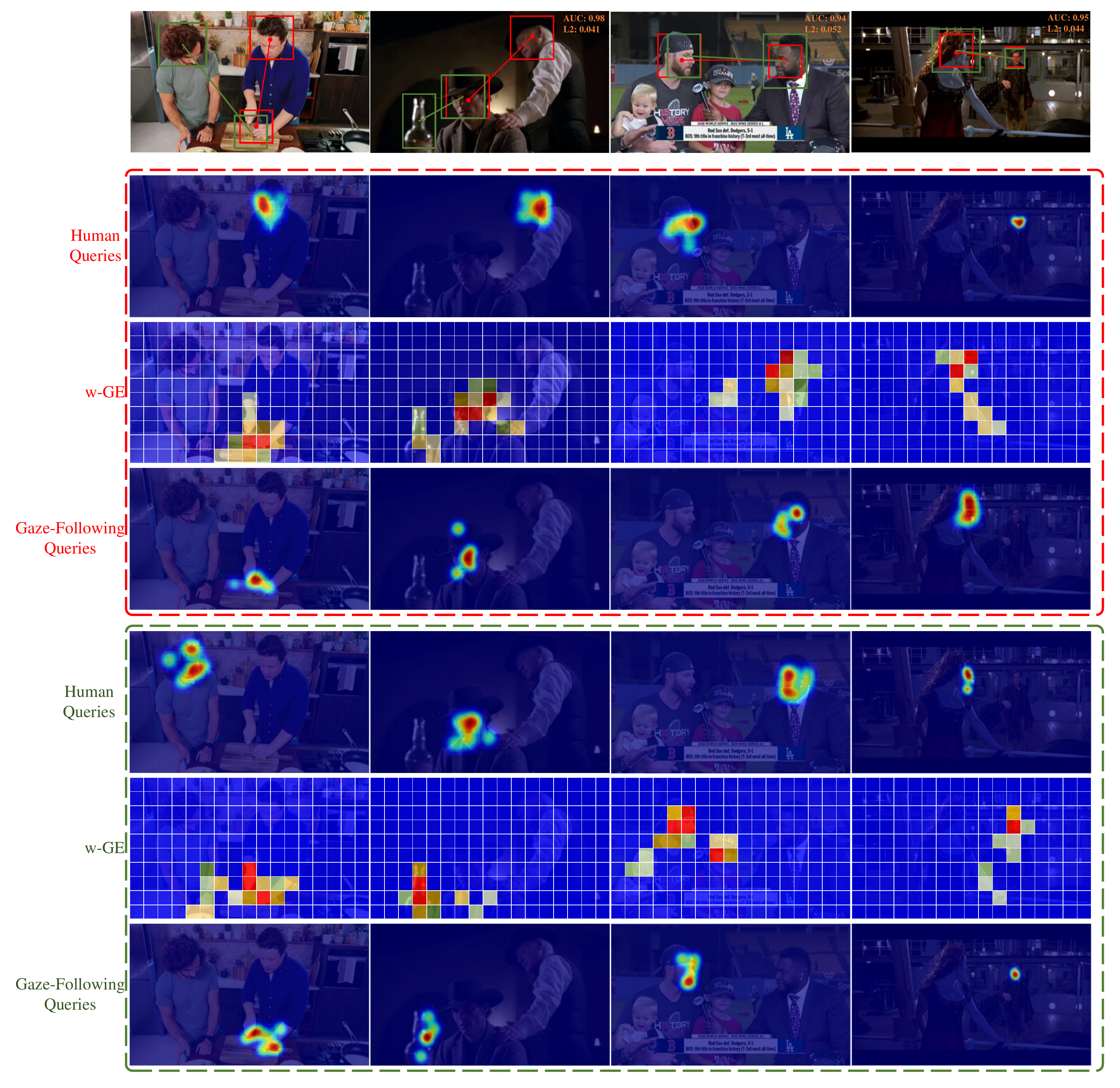}
        \captionsetup{size=small}
		\caption{{\bfseries Visualization} of human queries (row-II and row-V), w-GE (row-III and row-VI) and gaze following queries (row-IV and row row-VII) in our two-branch decoders. The images are randomly selected from the testing set of VideoAttentionTarget. Attention {\color{blue}{scores}} are coded with different colors for different humans. Best viewed in color.} 
		\label{attention}
  \vspace{-0.5em}
	\end{figure*}

\begin{figure*}[!t]
		\centering
		\includegraphics[width=\linewidth]{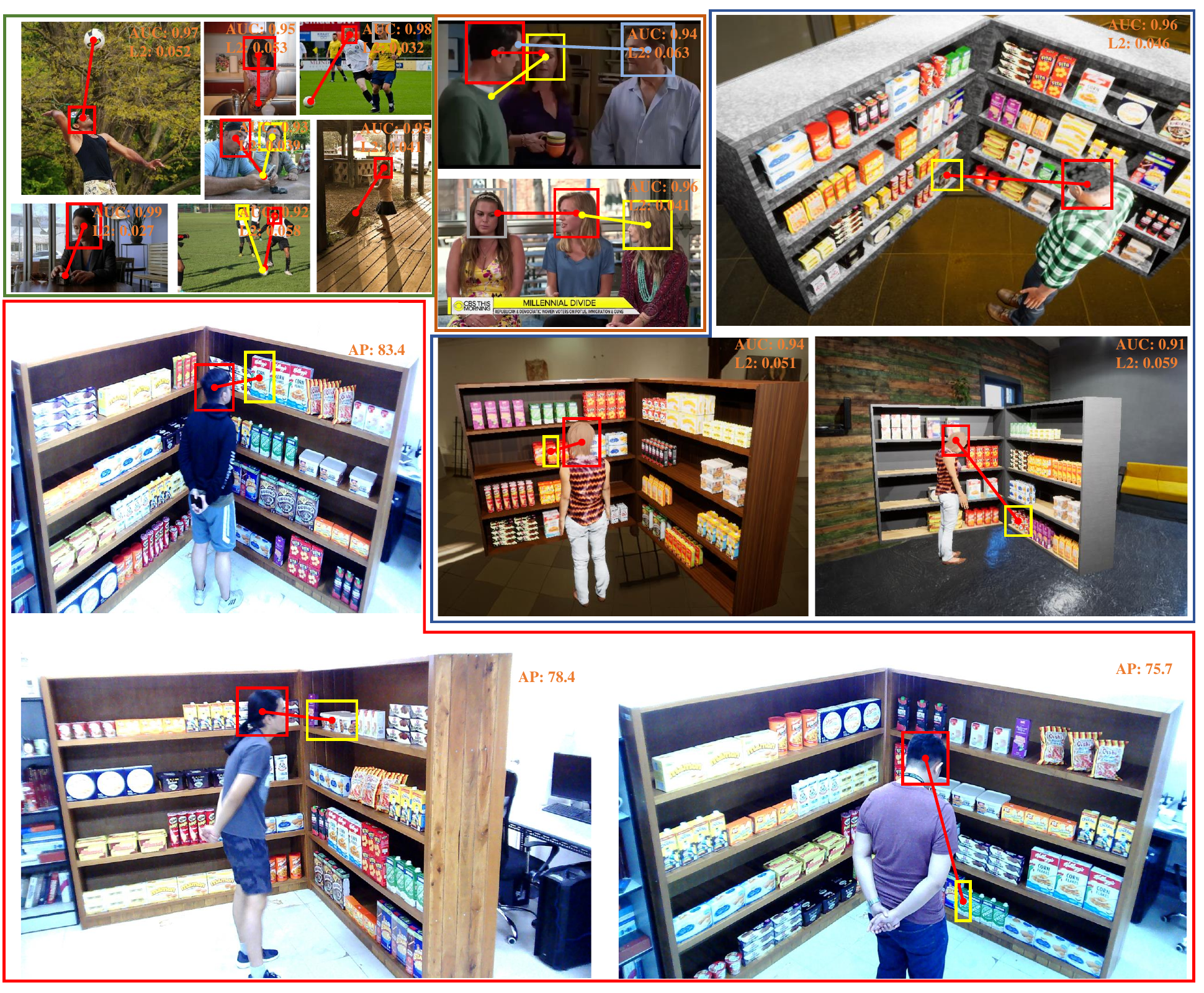}
        \captionsetup{size=small}
		\caption{{\bfseries Visualization} of detection results of GTR. The images are selected from GazeFollowing (in {\color{green}{green}} rectangle), VideoAttentionTarget (in {\color{brown}{brown}} rectangle), GOO-Synth (in {\color{blue}{blue}} rectangle) and GOO-Real (in {\color{red}{red}} rectangle).} 
		\label{more_case}
  \vspace{-0.5em}
	\end{figure*}

\vspace{0.1mm}
\noindent
\emph{2) Weight guided embedding}. Jointly Learned with human queries (\emph{j}-learned in Table~\ref{tab:wge})  through human decoder, w-GEs are proposed to highlight the potentially salient regions attended by different human queries. As shown in Table~\ref{tab:wge}, W-GE mechanism brings a $18.5\%$ AUC gain for GL-D and $39.1\%$ AP for GO-D. Interestingly, if we adopt an individual sub-branch to learn w-GE separately (\emph{s}-learned), the performances of GL-D and GO-D degrade $4.7\%$ and $10.8\%$, respectively. We conjecture that jointly learning enables an explicit information propagation from human queries to w-GE, thanks to the attention mechanism. We further qualitatively visualize the w-GEs in Fig~\ref{attention} (the third and the sixth rows), which shows that w-GE roughly highlight the salient regions that potentially be noticed by given human, guiding the subsequent gaze following detection. More examples are illustrated in Fig.~\ref{more_case}.

\begin{table*}[t]
    \centering
    \subfloat[\textbf{Position encoding}.
    \label{tab:pe}
    ]{
    \centering
    \begin{minipage}{0.24\linewidth}{\begin{center}
    \tablestyle{4pt}{1.3}
    \begin{tabular}{y{25}x{21}x{21}}
     & AUC & AP \\
    \shline
    none & 0.782 & 53.10 \\
    learned & 0.925 & \textbf{57.84} \\
    cosine & \baseline{\textbf{0.927}} & \baseline{57.12} \\
    \end{tabular}
    \end{center}
    }\end{minipage}
    }
    \subfloat[\textbf{Visual encoder}.
    \label{tab:ve}
    ]{
    \centering
    \begin{minipage}{0.24\linewidth}{\begin{center}
    \tablestyle{4pt}{1.3}
    \begin{tabular}{x{20}x{20}x{20}x{20}}
    CNN & T-E & AUC & AP \\
    \shline
    \cmark &  & 0.684 & 34.22 \\
     & \cmark & 0.751 & 42.64 \\
     \cmark & \cmark & \baseline{\textbf{0.927}} & \baseline{\textbf{57.12}}\\
    \end{tabular}
    \end{center}
    }\end{minipage}
    }
    \subfloat[\textbf{Decoder structure}.
    \label{tab:ds}
    ]{
    \centering
    \begin{minipage}{0.24\linewidth}{\begin{center}
    \tablestyle{4pt}{1.3}
    \begin{tabular}{y{30}x{20}x{20}}
     & AUC & AP \\
    \shline
    single & 0.907 & 47.17 \\
    \emph{s}-dual & 0.863 & 48.64 \\
    \emph{j}-dual & \baseline{\textbf{0.927}} & \baseline{\textbf{57.12}} \\
    \end{tabular}
    \end{center}
    }\end{minipage}
    }
   \subfloat[\textbf{Weight guided embedding}.
    \label{tab:wge}
    ]{
    \centering
    \begin{minipage}{0.24\linewidth}{\begin{center}
    \tablestyle{4pt}{1.3}
    \begin{tabular}{y{33}x{20}x{20}}
     & AUC & AP \\
    \shline
    none & 0.841 & 39.62 \\
    \emph{s}-learned & 0.885 & 52.44\\
    \emph{j}-learned & \baseline{\textbf{0.927}} & \baseline{\textbf{57.12}} 
    \end{tabular}
    \end{center}
    }\end{minipage}
    }

    \vspace{1.25mm}
    \subfloat[\textbf{Connection of decoders}.
    \label{tab:cd}
    ]{
    \centering
    \begin{minipage}{0.3\linewidth}{\begin{center}
    \tablestyle{4pt}{1.3}
    \begin{tabular}{y{40}x{25}x{25}}
     & AUC & AP \\
    \shline
    last-to-last  & 0.893 & 50.42\\
    last-to-all & 0.905 & 53.24\\
    one-to-one & \baseline{\textbf{0.927}} & \baseline{\textbf{57.12}}
    \end{tabular}
    \end{center}
    }\end{minipage}
    }
    \subfloat[\textbf{Gaze following queries}.
    \label{tab:gfq}
    ]{
    \centering
    \begin{minipage}{0.3\linewidth}{\begin{center}
    \tablestyle{4pt}{1.3}
    \begin{tabular}{x{25}x{25}x{20}x{20}}
      $\mathbf{Q}_h$ &  w-GE & AUC & AP \\
    \shline
    \cmark & & 0.841 & 39.62 \\
    & \cmark & 0.863 & 44.83 \\
    \cmark & \cmark & \baseline{\textbf{0.927}} & \baseline{\textbf{57.12}} \\
    \end{tabular}
    \end{center}
    }\end{minipage}
    }
    \subfloat[\textbf{Model depths}.
    \label{tab:md}
    ]{
    \centering
    \begin{minipage}{0.37\linewidth}{\begin{center}
    \tablestyle{4pt}{1.3}
    \begin{tabular}{x{20}x{20}x{30}x{22}x{22}x{22}}
    H & G & Param. & FPS & AUC & AP \\
    \shline
    3 & 3 & \baseline{48.2M} & \baseline{\textbf{27}} & \baseline{\textbf{0.927}} & \baseline{57.12}\\
    3 & 5 & 49.6M & 24 & 0.925 & 58.25 \\
    5 & 5 & 56.4M & 20 & 0.911 & \textbf{58.43}
    \end{tabular}
    \end{center}
    }\end{minipage}
    
    }
\captionsetup{size=small}
   \caption{\textbf{GTR ablation experiments} with ResNet-50 on VideoAttentionTarget (AUC for GL-D) and GOO-Real (AP for GO-D). We roundly verify our proposed strategies for model design and compare different possible variants. Default settings are marked in {\colorbox{baselinecolor}{gray}}.}
\end{table*}
\begin{table*}[t]

    \centering
    \subfloat[\textbf{Human head prediction cost}.
    \label{tab:h}
    ]{
    \centering
    \begin{minipage}{0.31\linewidth}{\begin{center}
    \tablestyle{6pt}{1.3}
    \begin{tabular}{x{20}x{20}x{20}x{30}}
    $\alpha_h^1$ & $\alpha_h^2$ & AP & Recall \\
    \shline
    1 & 1 & 91.2 & 73.6 \\
    1 & 2 & 89.4 & 70.8 \\
    2 & 1 &\baseline{\textbf{94.3}} & \baseline{\textbf{73.9}} \\
    \end{tabular}
    \end{center}
    }\end{minipage}
    }
    \subfloat[\textbf{gaze location prediction cost}.
    \label{tab:gl}
    ]{
    \centering
    \begin{minipage}{0.31\linewidth}{\begin{center}
    \tablestyle{4pt}{1.3}
    \begin{tabular}{x{20}x{20}x{20}x{30}}
    $\alpha_{gl}^1$ & $\alpha_{gl}^2$ & AUC &$L_2$ Dist. \\
    \shline
    1 & 1 & 0.912 & 0.114 \\
    1 & 2 & 0.905 & 0.117 \\
    2 & 1 &\baseline{\textbf{0.927}} & \baseline{\textbf{0.102}} \\
    \end{tabular}
    \end{center}
    }\end{minipage}
    }
    \subfloat[\textbf{Gaze object prediction cost}.
    \label{tab:go}
    ]{
    \centering
    \begin{minipage}{0.31\linewidth}{\begin{center}
    \tablestyle{4pt}{1.3}
    \begin{tabular}{x{20}x{20}x{20}x{30}}
    $\alpha_{go}^1$ & $\alpha_{go}^2$ & AP &$\text{AP}_{50}$  \\
    \shline
    1 & 1 & 56.40 & 96.18 \\
    1 & 2 & \baseline{\textbf{57.12}} & \baseline{\textbf{96.47}} \\
    2 & 1 & 55.83 & 96.02\\
    \end{tabular}
    \end{center}
    }\end{minipage}
    }

    \vspace{1.25mm}
    \subfloat[\textbf{Weights for bipartite matching}.
    \label{tab:ma}
    ]{
    \centering
    \begin{minipage}{0.48\linewidth}{\begin{center}
    \tablestyle{4pt}{1.3}
    \begin{tabular}{x{20}x{20}x{20}x{30}x{30}x{30}}
    $\sigma_1$ & $\sigma_2$ & $\sigma_3$ & AUC & AP & mAP   \\
    \shline
    1 & 1 & 1 & 0.922 & 57.14 & 0.625 \\
    2 & 1 & 1 & \baseline{\textbf{0.927}} & \baseline{57.12} & \baseline{\textbf{0.627}} \\
    1 & 2 & 2 & 0.914 & \textbf{57.92} & 0.623\\
    \hspace*{\fill}
    \end{tabular}
    \end{center}
    }\end{minipage}
    }
    \subfloat[\textbf{Weights for objective function}.
    \label{tab:of}
    ]{
    \centering
    \begin{minipage}{0.5\linewidth}{\begin{center}
    \tablestyle{4pt}{1.3}
    \begin{tabular}{x{20}x{20}x{20}x{20}x{30}x{30}x{30}}
    $\eta_1$ & $\eta_2$ & $\eta_3$ & $\eta_4$ & AUC & AP & mAP   \\
    \shline
    1 & 1 & 1 & 1 & 0.913 & 56.41 & 0.607 \\
    2.5 & 1 & 1 & 2 & \baseline{0.927} & \baseline{57.12} & \baseline{\textbf{0.627}} \\
    2 & 2 & 2 & 1 & \textbf{0.931} & 56.42 & 0.623\\
    1 & 1 & 1 & 2 & 0.923 & \textbf{58.04} & 0.622 
    \end{tabular}
    \end{center}
    }\end{minipage}
    }
\captionsetup{size=small}
   \caption{\textbf{Ablation experiments} for our new loss design, where we aim at verifying the mutual impacts of GL-D and GO-D in our unified framework. Meanwhile, we also explore the effects of regression and classification in different cost functions by changing their weights.}
\end{table*}
\begin{table}[t]
	\tablestyle{6pt}{1.2}
	\centering
 \captionsetup{size=small}
	\caption{{\bfseries GL-D in practical scenario}. ``Ang." refers to the angle between ground-truth gaze direction and the predicted one. }
	\begin{tabular}{y{67}x{25}x{35}x{25}x{25}}
		\toprule[0.7pt]
		Methods         & AUC$\uparrow$  & $L_2$ Dist.$\downarrow$  & Ang.$\downarrow$ & mAP$\uparrow$  \\
		\midrule
		 Gazefollow~\cite{recasens}          & 0.792           & 0.213           & $27.9^{\circ}$             & 0.407                         \\
		 Lian~\cite{lian2018}                & 0.813           & 0.167           & $19.7^{\circ}$             & 0.441                          \\
		 VideoAttention~\cite{chong}         & 0.842           & 0.145           & $16.9^{\circ}$             & 0.476                           \\
       GaTector~\cite{wang2022gatector}         & 0.837           & 0.152           & $17.3^{\circ}$             & 0.474                           \\
		
		\midrule
		\rowcolor{baselinecolor}
		GTR                 & \textbf{0.938}           & \textbf{0.112}           & $\mathbf{11.8^{\circ}}$            & \textbf{0.562}                           \\
		 	
	\bottomrule[0.7pt]
		
	\end{tabular}
	\label{tab:prat}
\end{table}
\begin{table}[t]
	\tablestyle{6pt}{1.2}
	\centering
 \captionsetup{size=small}
	\caption{{\bfseries Shared attention detection results on the VideoCoAtt dataset}. The interval detection is evaluated in terms of prediction accuracy while the location task is measured with $L_2$ distance.}
	\begin{tabular}{y{97}x{40}x{40}}
		\toprule[0.7pt]
		Methods         & Accuracy$\uparrow$  & $L_2$ Dist.$\downarrow$    \\
		\midrule
		 Random                                        & 50.8           & 286                                \\
		 Fixed bias                                    & 52.4           & 122                                 \\
		 GazeFollow~\cite{recasens}                    & 58.7           & 102                                  \\
		 Gaze+Saliency~\cite{pan2016shallow}           & 59.4           & 83                                    \\
		 Gaze+Saliency+LSTM~\cite{hochreiter1997long}  & 66.2           & 71                                     \\
		 Fan~\cite{fan2018inferring}                   & 71.4           & 62                                      \\
		 Sumer~\cite{sumer2020attention}               & 78.1           & 63                                       \\
		 VideoAttention~\cite{chong}                   & 83.3           & 57                                        \\
		
		\midrule
		\rowcolor{baselinecolor}
		GTR                 & \textbf{90.4}           & \textbf{41}                               \\
		 	
	\bottomrule[0.7pt]
		
	\end{tabular}
	\label{tab:share}
\end{table}

 \begin{figure*}[!t]
		\centering
		\includegraphics[width=0.96\linewidth]{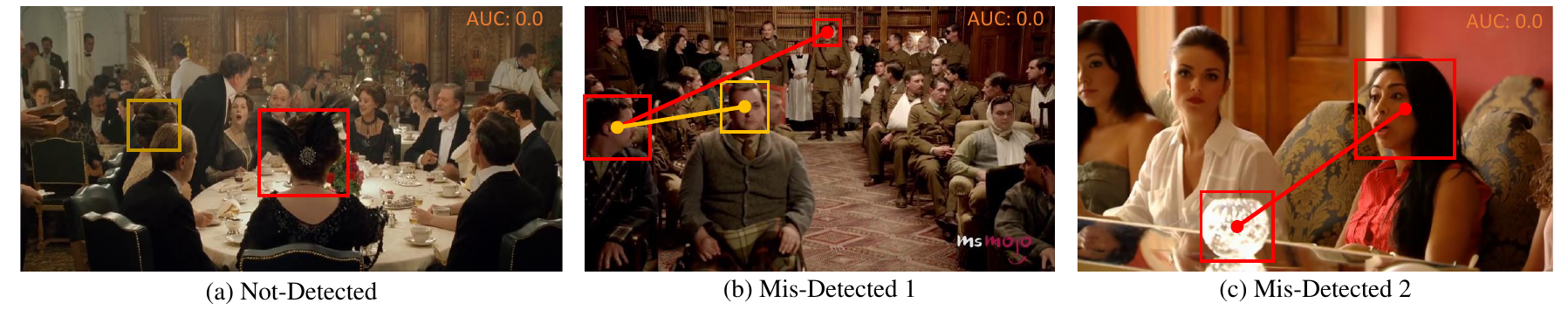}
        \captionsetup{size=small}
		\caption{{\bfseries Failed cases.} In (a), GTR fails to detect the human head due to the lack of discriminative facial features (the {\color{red}{red}} one) and occlusion (the {\color{yellow}{yellow}} one). In (b), the gaze location are mis-predicted (the {\color{red}{red}} one is the ground-truth) due to the salient object in the foreground (the {\color{yellow}{yellow}} one). In (c), the salient object misguides the prediction for the human is looking outsize of the image. The images are randomly selected from the testing set of VideoAttentionTarget and GazeFollowing.}
		\label{fail}
	\end{figure*}

\vspace{0.1mm}
\noindent
\emph{3) Gaze following decoder}. Our gaze following decoder is designed to iteratively interact with human decoder through a hierarchical architecture. First, if we cut off the connection between human decoder and gaze following decoder (\emph{s}-dual in Table~\ref{tab:ds}), the AUC for GL-D degrades $6.9\% \:(\nabla 0.064)$ and the AP for GO-D reduces $14.8\% \: (\nabla 8.48)$. Then, if we adopt an one-shot strategy, \emph{i.e.}, only connecting the last layer of human decoder with the last layer of gaze following decoder, the AUC and AP degrade $3.8\% \:(\nabla 0.034)$  and $13.3\% \: (\nabla 5.7)$, respectively, as shown in Table~\ref{tab:cd}. Another intuitive choice is to connect the last layer of human decoder (containing higher-lever semantic) with all layers of gaze following decoder, which yet leads a slight performance degradation ($\nabla2.4\%$ in AUC and $\nabla6.8\%$ in AP). We hypothesize that the feature from the last layer of human decoder has a semantic gap between the relative shallow layers in the gaze following decoder. Moreover, as shown in Fig.~\ref{fig:nums},  the human decoder is connected with the gaze-following decoder in a hierarchical and iterative manner, which enables the gaze-following decoder to filter out irrelevant objects gradually. Besides, unlike the human queries are randomly initialized, we explicitly initialize gaze following queries for each layer of gaze following decoder using the human queries and w-GE from the corresponding human decoder layer, whose effects are validated in Table~\ref{tab:gfq}. Intuitively, human queries provide human features while w-GEs introduce the context of potential salient regions, facilitating the learning of cross-attention weights between gaze following queries with global visual features.

\vspace{0.1mm}
\noindent
\emph{3) Model depth}. Empirically, increasing the depth of decoder, \emph{i.e.}, stacking more layers, is likely to improve performance at the cost of more computational costs. As illustrated by Fig~\ref{fig:dec}, a 3-layers decoder enables the optimal result for GL-D and that for GO-D is achieved by a 5-layers gaze following decoder. It is because that GL-D is a relative easier task than GO-D. We further try several different combines in depths of two-branch decoders, and find a 3-layer human decoder equipped with a 3-layer gaze following decoder is the optimal solution considering the trade-off in effectiveness and efficiency, as reported in Table~\ref{tab:md}.

\subsection{Ablation Study for Loss Design}
We also conduct extensive ablation experiments to analysis the effectiveness of our proposed new loss function for gaze following detection. We find that for human head detection and GL-D, regression plays a more important role than classification, which are yet reverse for GO-D (Table~\ref{tab:h},~\ref{tab:gl},~\ref{tab:go}). Specially, the ``AP" and ``Recall" in Table~\ref{tab:h} are metrics to evaluate the performance of human head detection, where a human head prediction is considered as true positive only if it localizes the human head with an IOU ratio between the predicted head box and the ground-truth one is larger than 0.7. We further compare several different weights for different costs in matching loss and objective loss, the results are reported in Table~\ref{tab:ma} and Table~\ref{tab:of}, respectively. Actually, for close to the last decade, gaze following detection has always been dominated by distance-based loss, such as $L_1$ and $L_2$ loss. We hope our new loss can provide some desirable points for gaze following community.

 \subsection{Generalization Performance}
 
\vspace{1.25mm}
\noindent
\textbf{Practical application}. To evaluate the GTR's generalization performance in practical scenarios, we randomly select 15,000 image from DL Gaze dataset~\cite{lian2018}, which records the daily activities of 16 volunteers in 4 diverse scenes and the ground-truth annotations are labeled by the observers in the videos. We compare the performance of different models by training them just with the VideoAttentionTarget dataset and then applying them for DL Gaze without fine-tuning. As the quantitative results reported in Table~\ref{tab:prat}, our GTR outperforms all state-of-the-art methods by a significant margin, demonstrating its impressive performance in practical applications.

\vspace{1.25mm}
\noindent
\textbf{Shared attention detection}. With the ability to detect gaze locations of different people simultaneously, GTR are natively able to understanding shared attention in social scenes. We report the results in terms of accuracy for interval detection of shared attention and $L_2$ distance for location prediction on the VideoCoAtt~\cite{fan2018inferring} by following~\cite{chong}. VideoCoAtt contains 113,810 tset frames that are annotated with the target location when it is simultaneously attended by two or more people. As can be seen from Table~\ref{tab:share}, GTR shows a excellent potntial value of recognizing higher-lever social gaze.

\section{Discussion}
\label{sec:dis}
Transformer demonstrates impressive capability in gaze following detection. We conjecture the reasons are twofold: 1) Long range context modeling is essential in gaze following detection. Compared to CNN with limited receptive field, the global attention mechanism ensures Transformer to capture holistic scene feature, providing crucial cues to reason possible gaze location. 2) Interactions between holistic scene contexts and head pose features are important. We reframe gaze following detection task as detecting human head locations and their gaze followings simultaneously, which implicitly enforces the model to learn the interactions between these two different features. Besides, we design the decoder of our GTR as two parallel branches, one of which aims for head detection and another one is responsible for gaze following detection. Meanwhile, these two branch is connected in a hierarchical and iterative manner, which explicitly enables a strong interaction between holistic scene contexts and head features. Besides the model design, it is our belief that defining gaze following detection task as simultaneously detecting human head locations and their gaze following is more reasonable, since manually labeled head locations are unavailable in practical scenarios. Moreover, it allows us to detect multiple humans' gaze followings at once, saving computational complexity and being more efficient.

However, GTR sometimes may be failed, and we exemplify some representative failed cases in Fig. 5. In some scenarios, such as the human head is partly occluded or he/she turns his/her back to the camera, the GTR may fails to localize the human head due to the missing of discriminative facial features. Actually, occlusion has always plagued detection system and many excellent methods~\cite{zhou2019occlusion, kim2019bbc} have been proposed to solve it, among which data augmentation may be one of effective solutions for the GTR. In terms of gaze following detection, GTR sometimes may be deluded by the foreground objects since they are likely to be more salient. In this case, using additional cues, \emph{e.g.}, depths~\cite{fang2021dual}, can be a great choice to guide the GTR.

\section{Conclusion}
\label{sec:con}
In this paper, we present a new Transformer-based method for the task of gaze following detection, named as GTR. GTR models global contextual information, being designed to localize all humans' head crops as well as to detect their gaze following simultaneously. GTR unites gaze location detection and gaze object detection in a unified framework, which is also the first single-stage pipeline where the model can be trained in a fully end-to-end manner. Extensive experiments validate its impressive performance in both effectiveness and efficiency. Besides, GTR shows desirable potential for understanding gaze behavior in naturalistic human interactions. We hope our method will be useful for human activity understanding research.

{\small
	\bibliographystyle{ieee_fullname}
	\bibliography{egbib}
}

\end{document}